\documentclass[12pt,letterpaper]{article}
\usepackage[top=0.6in,left=1.5in,footskip=0.75in,marginparwidth=2in]{geometry}
\date{}

\usepackage{moreverb,url}

\usepackage[pdftex]{graphicx}
\usepackage{subfigure}

\usepackage{algorithmic}
\usepackage{algorithm}
\usepackage{amssymb,amsmath}
\usepackage{stmaryrd}
\usepackage{xcolor}
\usepackage{multicol}
\usepackage{multirow}
\usepackage[T1]{fontenc}
\usepackage{xintexpr}
\usepackage{leftidx}

\usepackage{rotating}
\usepackage{multirow}
\usepackage{longtable}
\usepackage{colortbl}
\usepackage{dsfont}

\usepackage{pgf}

\usepackage{url}
\DeclareMathOperator*{\argmax}{arg\,max}

\newcommand{\ue}{%
	\begin{tikzpicture}%
	\draw[fill = black] (.25ex,.25ex) circle (.3ex);
	\draw[thick] (.55ex,.25ex) -- (1.55ex,.25ex);%
	\draw[fill = black] (1.85ex, .25ex) circle (.3ex);%
	\end{tikzpicture}%
}

\newcommand{\simpcir}{%
	\begin{tikzpicture}%
	\draw[fill = black] (.25ex,.25ex) circle (.3ex);
	\end{tikzpicture}%
}

\usepackage{tikz}
\usepackage{centernot}%
\usetikzlibrary{shapes.geometric}

\tikzstyle{extremely densely dashed} = [dash pattern=on 2pt off 1pt]
\tikzstyle{pattern}=[draw,circle,black,bottom color=white, top color= white, text=black,minimum width=10pt]
\tikzstyle{peers}=[draw,circle,violet,bottom color=green!60, top color= white, text=violet,minimum width=10pt, scale = 0.7]
\definecolor{burntorange}{cmyk}{0,0.52,1,0}
\def\oran{orange!30}
\tikzstyle{cliquepeers}=[draw, red, fill= red, text=black,minimum width=1pt, scale=0.4]
\tikzstyle{myFivePoly} =  [regular polygon,regular polygon sides=5,minimum width=1pt, scale=0.9]
\tikzstyle{mytriangle} =  [fill=blue!20, regular polygon, regular polygon sides=3,minimum width=1pt, scale=0.4]
\tikzstyle{extremely densely dashed} = [dash pattern=on 2pt off 1pt]
\tikzstyle{finedashpath}=[violet, extremely densely dashed,thick]
\tikzstyle{superpeers}=[draw,circle,thick,burntorange, left color=\oran,text=black, minimum width=20pt]
\tikzstyle{pattern}=[draw,circle,black,bottom color=white, top color= white, text=black,minimum width=10pt]

\newcommand{\samplePattern}{%
	\begin{tikzpicture}%
	\node[pattern, inner sep=0pt, bottom color = gray, scale=0.5] (samplePattern) at (0,0) {};
	\end{tikzpicture}%
}

\newcommand{\inlineimage}[1]{$\vcenter{\hbox{\protect\includegraphics[height=0.8\baselineskip,origin=c]{#1}}}$}

\begin{document}

\title{Unsupervised Learning of Shape Concepts -- \\ From Real-World Objects to Mental Simulation}

\author{Christian A. Mueller~and~Andreas Birk
\thanks{The authors are with the Robotics Group of the Computer Science \& Electrical Engineering Department, Jacobs University Bremen, Germany \newline e-mail: \{chr.mueller, a.birk\}@jacobs-university.de.}%
	}

\markboth{IEEE Transactions on Cognitive and Developmental Systems,~Vol.~XX, No.~X, X~XXXX}%
{Mueller and Birk: Machine-Centric Conceptualization of Shape Commonalities Through Mental Simulation of Artificially Generated Objects}

\maketitle

\begin{abstract}
  \small
  An unsupervised shape analysis is proposed to learn concepts reflecting shape commonalities. 
  Our approach is two-fold: i) a spatial topology analysis of point cloud segment constellations within objects is used in which constellations are decomposed and described in a hierarchical and symbolic manner. 
  ii) A topology analysis of the description space is used in which segment decompositions are exposed in. 
  Inspired by Persistent Homology, groups of shape commonality are revealed. 
  Experiments show that extracted persistent commonality groups can feature semantically meaningful shape concepts; the generalization of the proposed approach is evaluated by different real-world datasets.   
  We extend this by not only learning shape concepts using real-world data, but by also using mental simulation of artificial abstract objects for training purposes. 
  This extended approach is unsupervised in two respects: \emph{label-agnostic} (no label information is used) and \emph{instance-agnostic} (no instances preselected by human supervision are used for training).  
  Experiments show that concepts generated with mental simulation, generalize and discriminate real object observations.  
  Consequently, a robot may train and learn its own internal representation of concepts regarding shape appearance in a self-driven and machine-centric manner while omitting the tedious process of supervised dataset generation including the ambiguity in instance labeling and selection.
\end{abstract}

\section{Introduction and Motivation}
Studies of early object perception in infants~\cite{SPELKE199029} suggested that objects can be characterized by a set of properties such as continuity, i.e., objects successively move along a path, or solidity, i.e., objects can only move through free-space. 
Furthermore, shape is a key visual cue as it fundamentally contributes to reasoning and understanding of objects~\cite{doi:10.1111/1467-9280.03439,Graf2010}.
Inferred shape commonalities among objects allow to infer similar object (including semantic) properties. Shape is used in many robotic application areas ranging from household to industry, e.g., in object shape categorization tasks~\cite{6942984}, in generation of grasping primitives for similar object appearances in manipulation~\cite{6696928}, or in finding substitutes for currently absent objects~\cite{DBLP:conf/icra/AbelhaGS16,ThosarMuellerZugToolSubArXiv2018}, to name just a few examples.

In traditional object perception in form of object instance or category recognition, an association is formed between a label and a specific instance (e.g., \emph{John's mug}) or a generic group of instances (e.g., \emph{mug}), which share commonalities in appearance~\cite{Sloutsky2010}. 
A group of instances can be denoted as a \emph{category} and the description and abstraction of group commonalities as a \emph{concept}.
Learning concepts from objects by associating meaning to a system's percepts is often conducted through interaction \cite{7174991} and supervision \cite{8023995,7451222}.
Eventually, associations are generally human-made, individually and continuously evolved over lifetime experience~\cite{Ashby1999} based on a set of modalities like tactual, auditory or visual sensations~\cite{Seward97a,Palmeri2004a}.
The combination of those sensations allows us to reliably interpret perceived object information~\cite{ZimgrodHommel2013}.
Humans are capable of incorporating further modalities including functional object knowledge to differentiate even though visual percepts can be similar, e.g., \emph{mug}, \emph{cup}, \emph{vase} or \emph{bowl}.
Consequently, such natural concepts are often not inferable from a machine-perspective due to the lack of dimensionality representing the perceived observations (e.g., only images or point clouds).
From a machine-vision perspective, human-supervised learning methods are particularly highly vulnerable to incorporate such knowledge, e.g., the function or affordances of objects, which is not inferable from pure sensor data. 
This is often inevitable when a supervised labeling process is conducted by humans, which will ultimately lead to biases in the learning phase.
\begin{figure}[tb]
	\small
	\centering
	\includegraphics[width=0.99\linewidth]{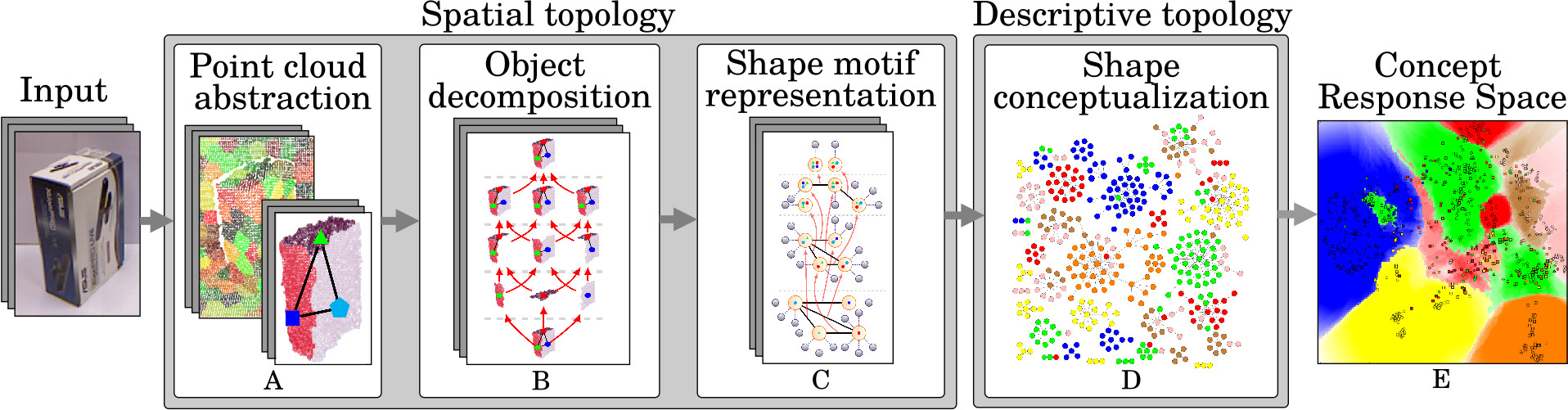}
	\caption{Illustration of the proposed object shape conceptualization approach.}
	\label{fig:approach_illustration}
\end{figure}

Our work in contrast focuses on object understanding from a machine-perspective avoiding supervision.
The work presented here builds upon our method from \cite{mueller_birk_iros2018} that learns shape concepts in an unsupervised (label agnostic) and data-driven manner from point clouds irrespective of human-annotations, which may contain biases.
Extracted segment constellations within object point clouds are used to learn patterns and eventually concepts of shape commonalities in a hierarchical manner.
It is shown that concepts can be learned from real-world RGBD-snapshots of objects, or more precisely single view point clouds omitting the color information, using well-known datasets like the \emph{Washington RGB-D Object Dataset} \cite{5980382} (WD) or the \emph{Object Segmentation Database} \cite{6385661} (SD). 
From the machine perspective, the concepts learned are purely derived by the given data, i.e., they are not affected by variable biases that may be caused by individual human interpretations with respect to the instance label annotation. As can be seen, the concepts learned on one real-world dataset generalize well across other real-world datasets not seen before.

Nevertheless, biases can be in the selection of the dataset instances used for training and validation. 
Moreover, the dataset generation process is cumbersome and generally requires effort in preparation including object instances selection, defining the experimental setup or labeling of object ground truth with regard to the background.
Therefore, we further investigate in this article the capability of learning the essence of object appearance from artificially generated object observations in simulation and whether the learned concepts are applicable to discriminate real-world object concepts.
This capability allows an artificial system like a robot to train its own internal representation of concepts regarding shape appearance in a self-driven and machine-centric manner without human-bias.
Thus, we present here an approach, which is unsupervised in two respects: it is \emph{label-agnostic} (no label information is used) as well as \emph{instance-agnostic} (no instances preselected by human supervision are used).

\section{Approach and Related Work}\label{sec:rw}
Shape analysis relies on a robust description and representation~\cite{Biasotti:2008:DSG:1391729.1391731,dicarlo:tics_2007} of objects, particularly in real world scenarios where snapshots of objects are affected by sensor noise and occlusions~\cite{JonschkowskiEHM16}. 
Theories of object perception from Cognitive Science and Psychology  suggest a hierarchical and component-based representation of object information~\cite{Fodor1988}.
Inspired by this, an analysis of topological patterns is applied here to sensor data in form of point clouds observed from single viewpoints with a Kinect-like camera.
The analysis is two-fold (see Fig.~\ref{fig:approach_illustration}): \textbf{i)} an analysis of the spatial topology in point cloud decompositions, \textbf{ii)} a topology analysis of these decompositions in description space.

Regarding \textbf{i)}, point clouds are initially over-segmented~\cite{Papon13CVPR} 
(Fig.~\ref{fig:approach_illustration} \textbf{A}) and further post-processed to segments that can reflect meaningful components of objects.
Subsequently, a hierarchical decomposition of point clouds is generated in a bottom-up manner (Fig.~\ref{fig:approach_illustration} \textbf{B}).
These segment compositions of objects allow to reason about shape characteristics and commonalities; commonalities observed within objects can be generalized to a shape concept.

Constellation models, which learn concepts from perceived feature (e.g., keypoints or segments) constellations have been successfully used in recent years~\cite{6942984,local_ContextSemanticLabelling-IJRR13,Leibe04combinedobject,Prasad11a,7139358}.
The inference is typically based on \emph{local} analysis of feature coherences with a priori learned constellation models, i.e., local evidences in a constrained spatial range with respect to the features using, e.g., Markov Networks~\cite{citeulike:8742196,10.1109/TPAMI.1984.4767596}.
This inference is robust to the absence of features due to noise and partial object occlusion.
Shape facets, which become apparent on a \emph{global} scale -- especially in case of complex structures, are in contrast insufficiently reflected considering only local inferences.
In the work presented here, a hierarchical constellation model (Fig.~\ref{fig:approach_illustration} \textbf{C}) is proposed in which segment constellations are decomposed over multiple topological levels that gradually (from local to global) reflect shape facets: from individual segment occurrences over segment groups to a single group of segments, which represent an entire object.
On each topological level, shape characteristics are observed and learned.

A related research field focuses on compositional hierarchies~\cite{DBLP:conf/nips/Utans93,FidlerChapter09,DBLP:conf/iccv/OzayAWL15} in which general geometric building-blocks like edges or contours are hierarchically composed to unions of these building-blocks.
Similarly, skeletonization methods \cite{Biasotti:2008:DSG:1391729.1391731,8099896,8000414} try to extract structure within objects from which regions and object components can be decomposed for reasoning purposes.
Our work differs in several aspects; especially as here a) the building-blocks are represented as symbols which characterize underlying 3D point cloud segments, and b) their constellations are subsequently learned in a multi-hierarchical manner.

Regarding \textbf{ii)}, i.e., the topology analysis of the decompositions in description space:
observed decompositions over the topological levels are here analyzed to gather distinctive insights and patterns that can be interpreted and related to concepts of specific shape appearances.
Persistent Homology (PH) is a concept related to Topological Data Analysis that has been applied in various areas related to high dimensional data visualization or to finding relations and coherencies in Big Data scenarios in general \cite{carlsson2014}.
PH allows to extrapolate features from data by means of finding persistent (or stable) feature appearances through an iterative filtration of the data compared to standard clustering approaches.
Standard cluster algorithm (e.g., k-Means, Expectation-Maximization, tree-based algorithms, etc.) associate data points to groups of data, which share similar properties, which is measured by a metric or similarity function. 
Inherent parameters of clustering algorithms are related to the number, size, variance of clusters, neighborhood distance between data points or in case of tree-like clustering, a splitting criterion. 
The parameterization is often computationally costly and it depends on the concrete data on which the clustering process is applied to.
Furthermore, partitioning the topology of a continuous description space with a static parameterization is often not a good solution due to over- and under-fitting effects.
Soft-clustering approaches like probability-based Expectation-Maximization (EM) provide a feedback of the actual fit of a query to the set of previously extracted clusters; but such approaches require additional post-processing in order to make a final decision about cluster membership.

PH in contrast allows to investigate the topological evolution of the data in a step-wise manner.
The concept of PH has already shown its applicability in geometric shape analysis to detect persistent shape patterns when being directly applied on point cloud data~\cite{carlsson2014,6909654,7487710}.
But instead of directly applying PH on point cloud data, we use here the responses that are retrieved from our topological analysis of point cloud decompositions proposed in \textbf{i)}.
The PH-based analysis allows to detect persistent appearances of the responses during the filtration process, which reveal shape commonalities of instances that can form concepts (Fig.~\ref{fig:approach_illustration} \textbf{D}).

Consequently, we focus on shape reasoning with a symbolic representation of geometric information, which is further exploited to learn visual patterns from observed object point cloud compositions on multiple granularity levels which allow to learn concepts from.
Our final goal is to investigate whether the visual patterns can be learned in a data-driven manner by encoding real object observations or on the basis of abstract artificial data from simulation.
The question arises whether artificial data from simulation allows to encode visual patterns that lead to concepts which can then be applied to discriminate real object observations (Fig.~\ref{fig:approach_illustration} \textbf{E}).

\section{Spatial Topology Analysis}
\label{sec:spatial_topo_analysis}

\subsection{Object Segment Extraction}
\label{sec:obj_seg_dict}

\begin{figure}[htbp]
    \small
	\centering %
	\subfigure[Point cloud abstraction ]{\label{fig:lower_segmentation}\includegraphics[width=0.45\linewidth]{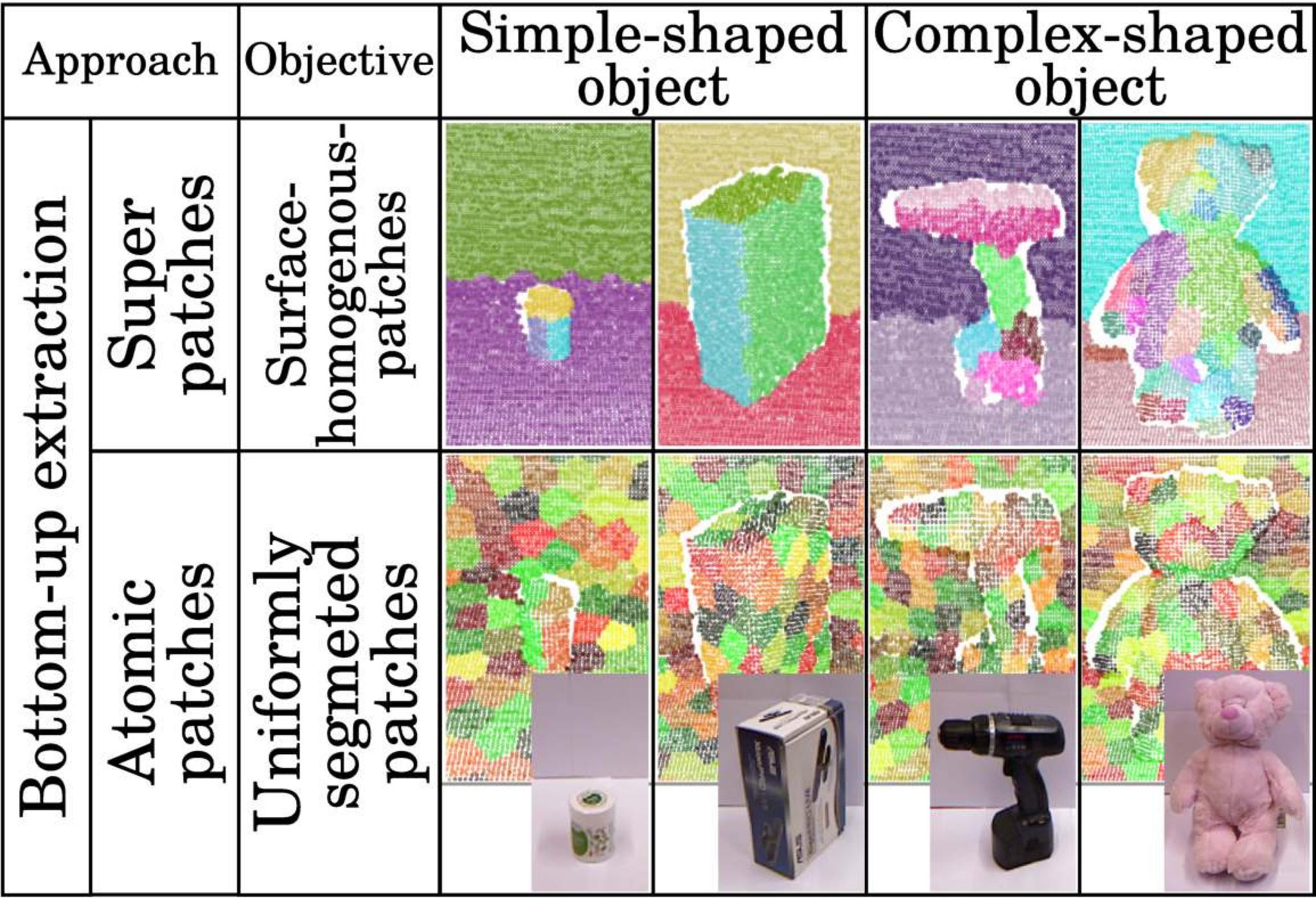}} \hfill
	\subfigure[Hierarchical dictionary]{\label{fig:dictionary}\includegraphics[width=0.53\linewidth]{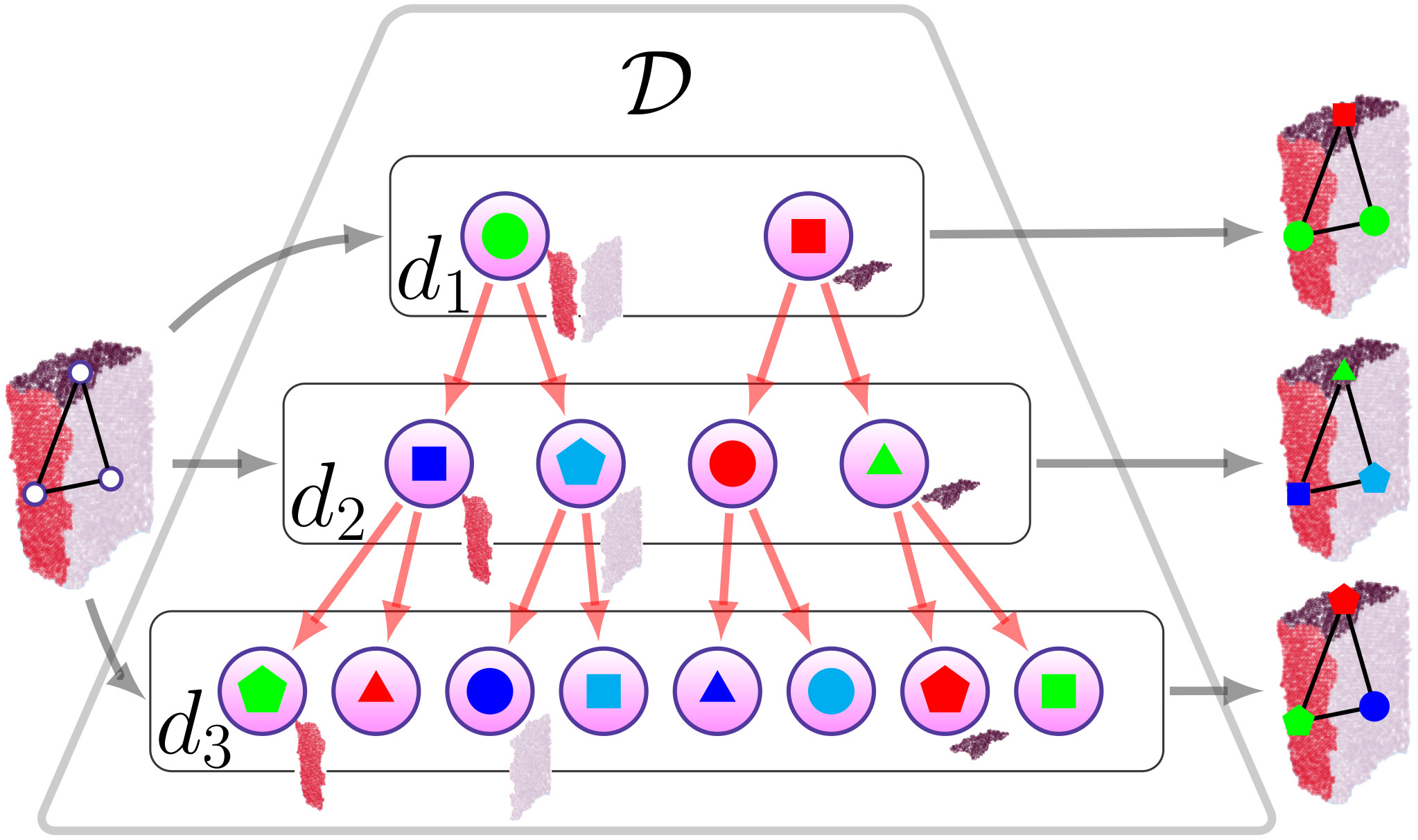}}
	\caption{\subref{fig:lower_segmentation} A two-step segmentation \cite{MuellerBirkIcra2016} from atomic patch segments to super patches is used as basis -- here illustrated by sample snapshots of 4 example objects  (\emph{can}, \emph{box}, \emph{cordless drill}, \emph{teddy}). \subref{fig:dictionary} An example hierarchical dictionary~\cite{6942984} $\mathcal{D}\mathrm{=}\{d_1,d_2,d_3\}$ is shown that consists of 3 description levels using divisive clustering. For illustration, each visual word is depicted as a circle with a colored polygon. A segmented object is shown as a graph on the left of the dictionary; on its right, the visual words assigned for each segment according to the respective description level are shown.} 
	\label{fig:eval:tsne_cfmat}
\end{figure}

Building on our work from  \cite{MuellerBirkIcra2016} as basis, an object point cloud is initially over-segmented into atomic patches and further processed to segments, also known as super patches, which can represent semantically meaningful shape components like planar surfaces of a \emph{box} or cylindrical and planar surfaces of a \emph{can} (see Fig.~\ref{fig:lower_segmentation}).
Subsequently, objects are represented as a set of \emph{point cloud segments}.
These segments can be interpreted as \emph{building blocks} that constitute objects.

Tackling with real world data, object observations are imperfect, e.g., noisy and partially occluded, which leads to a degradation of the detection of these building blocks.
This leads to failures in associating observed data to known building blocks, which is also in general known as the correspondence problem.
Therefore the stability of the detection of such building blocks in real world data is a major challenge.
To mitigate the correspondence problem among imperfect segments, a symbolic representation of segments is chosen as an abstraction step to facilitate further shape reasoning. 
Segment appearances are quantized to a set of discrete visual words following the well-known bag-of-words methodology \cite{6942984}, i.e., the visual words constitute a dictionary.
The idea is that similar appearing segments are abstracted to the same symbol, respectively, visual word.
The level of quantization plays a crucial role, since too few words may lead to under-fitting, whereas too many words may lead to over-fitting symptoms.
For an unbiased and purely data-driven word generation, a hierarchical divisive clustering procedure is applied as introduced in our previous work \cite{6942984}.
Therein segments are initially described with a description vector that is generated by a point cloud descriptor like FPFH \cite{5152473}.
As a result of the clustering procedure, a hierarchical dictionary $\mathcal{D}$ is created that consists of multiple description levels $\{d_1, d_2,...\}$, where level $f$ consists of $2^f$ words (Fig.~\ref{fig:dictionary}).
Each word represents a description vector whose position is inferred by the clustering procedure during the training phase using a set of segments captured from random scenes. 
Given an object segment, the extracted description vector of the segment is passed through the hierarchical dictionary $\mathcal{D}$. 
For each description level, the propagated description vector is accordingly labeled with the visual word whose description vector is closest using the $l^2$-norm (Fig.~\ref{fig:dictionary}). 

\subsection{Hierarchical Object Decomposition and Representation}
\label{sec:simu:desc_rep}
A segment composition of a captured object $o$ is initially represented as graph $g^o$ in which each segment corresponds to a vertex and neighboring vertices are connected with an edge.
Each vertex is augmented with the corresponding point cloud segment and the visual word that is inferred from the set of visual words on the respective description level in the dictionary $\mathcal{D}$ (see Sec.~\ref{sec:obj_seg_dict}); the visual word inferences can hence differ according to the description level as illustrated on the right side in Fig.~\ref{fig:dictionary}.

\begin{figure}[h!]
    \small
	\centering
	\begin{minipage}[b]{.39\linewidth}
		\subfigure[A shape motif hierarchy $\mathcal{H}$]{\label{fig:ch_illustration}\includegraphics[width=1.0\linewidth]{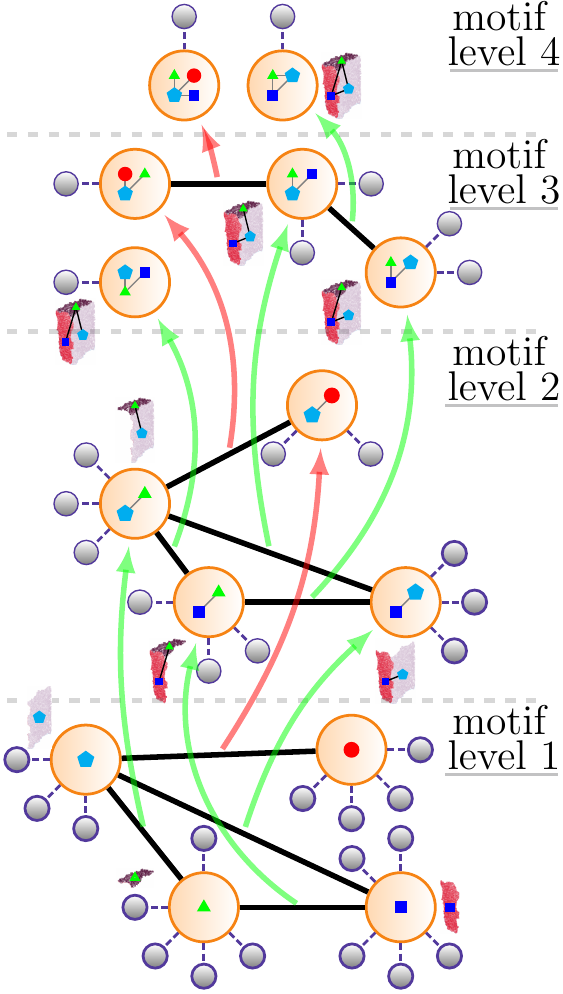}}
	\end{minipage} \hfill
	\begin{minipage}[b]{.59\linewidth}
		\centering
		\subfigure[A shape motif level]{\label{fig:ch_legend}\includegraphics[width=0.77\linewidth]{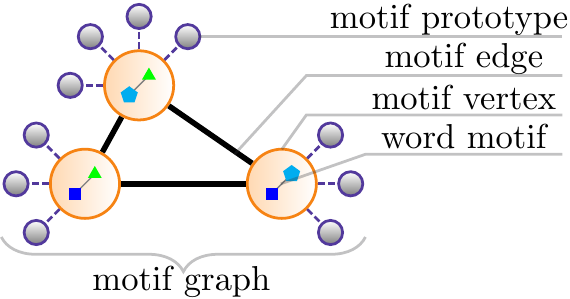}}
		
		\subfigure[A shape motif hierarchy ensemble]{\label{fig:hch_illustration}\includegraphics[width=1.0\linewidth]{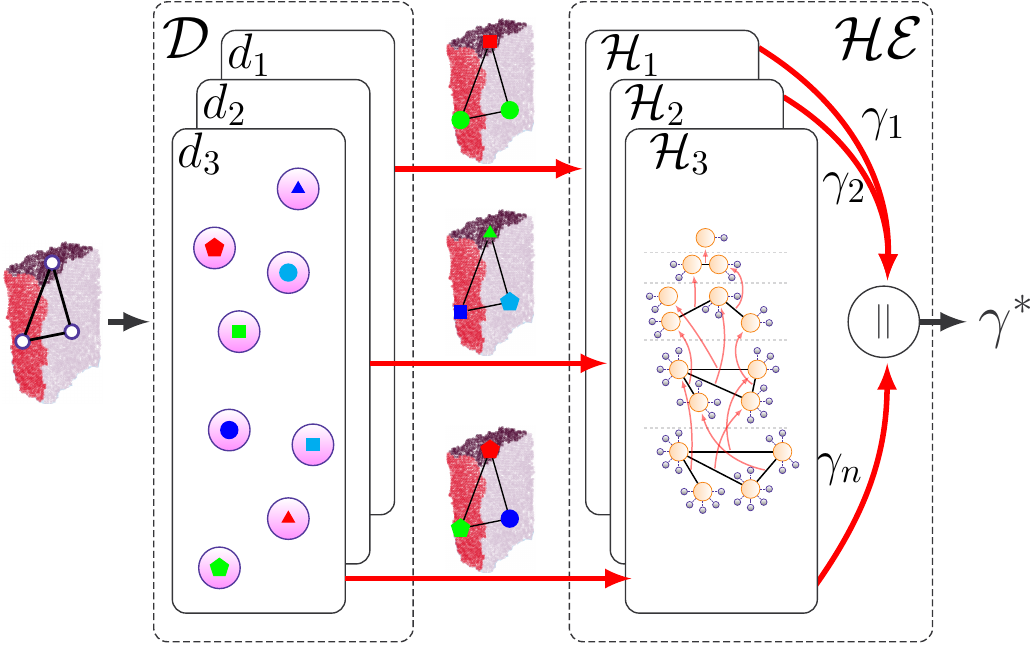}}
	\end{minipage}
	\caption{An example of a shape motif hierarchy $\mathcal{H}$ is shown in \subref{fig:ch_illustration}; it consists of multiple \emph{motif levels}. Each node \inlineimage{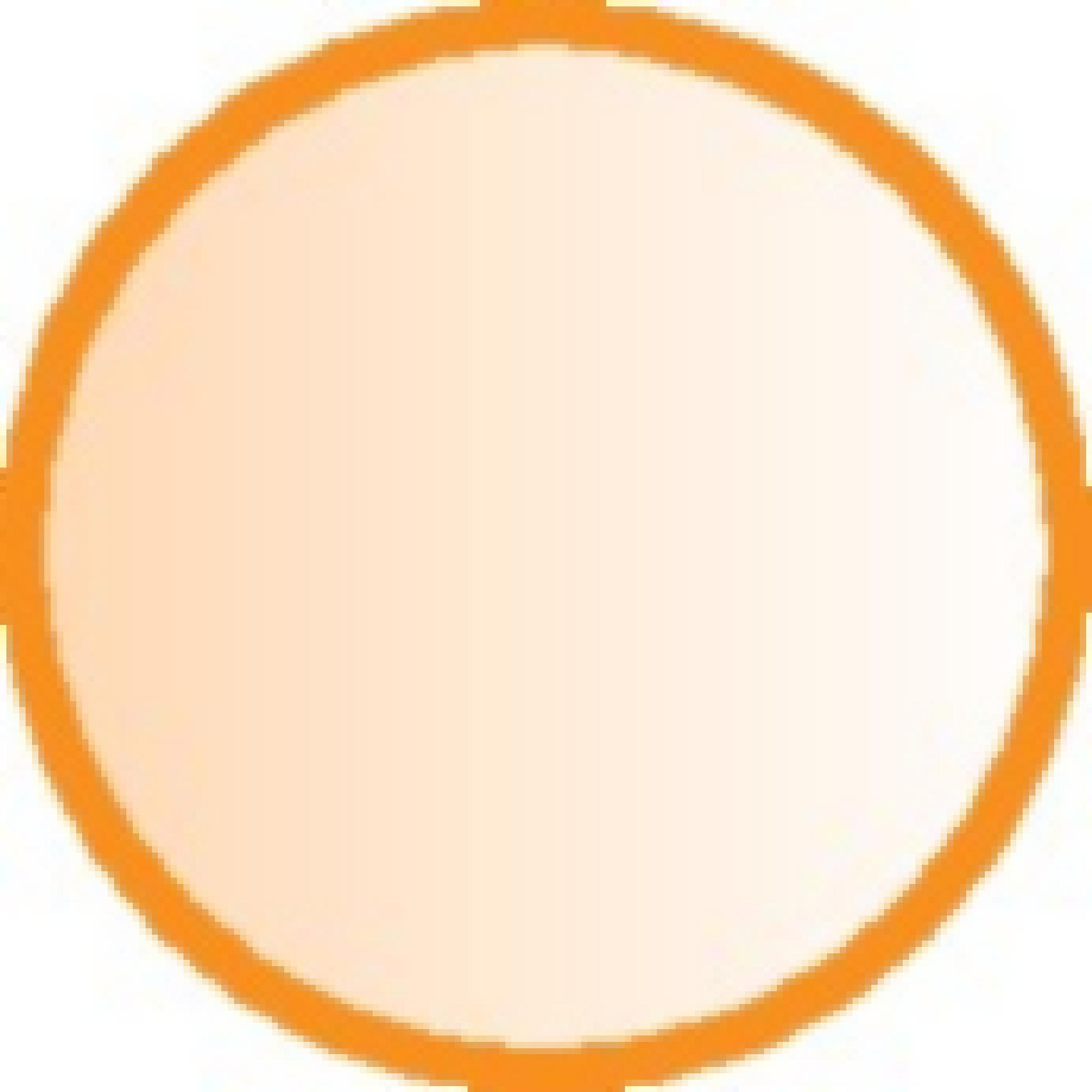} represents a specific \emph{motif vertex}, whereas each smaller linked node \protect \samplePattern\ represents a \emph{motif prototype}. A sample propagation (\protect \inlineimage{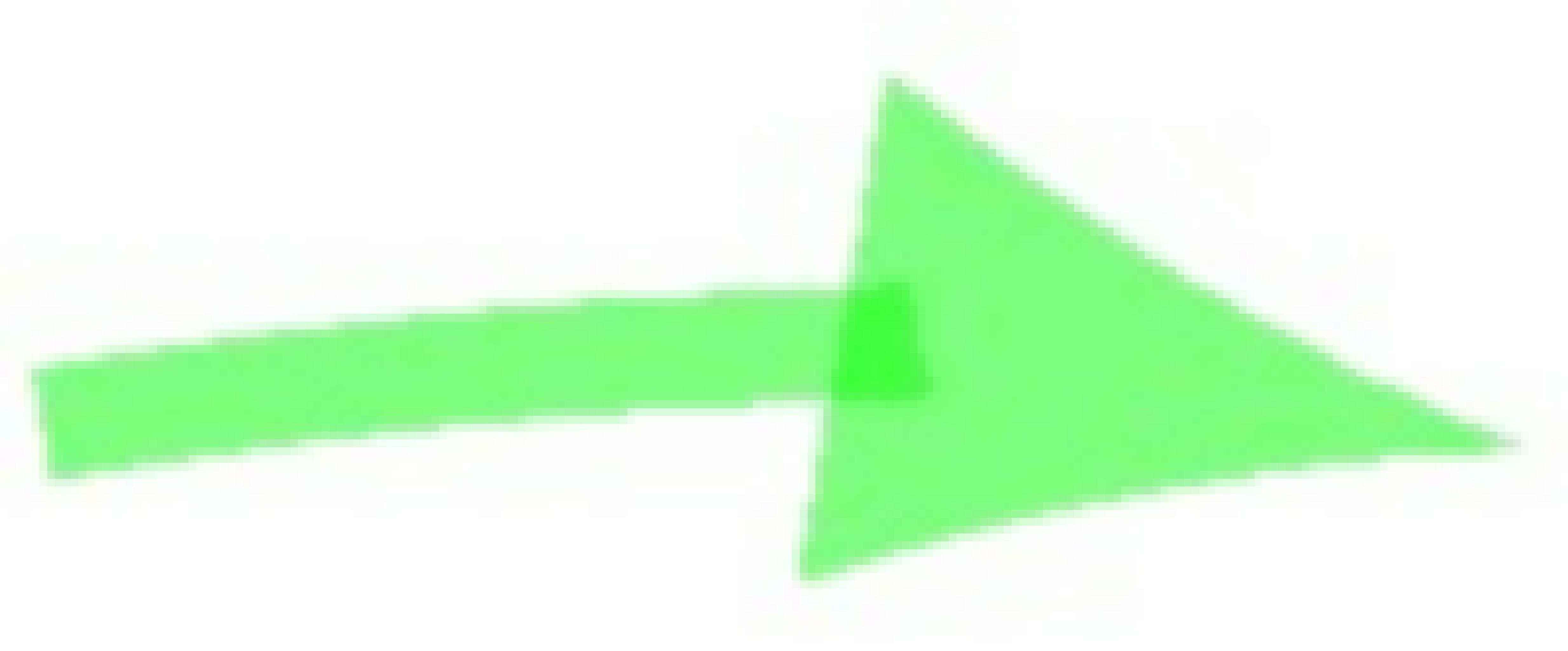}) of a box \inlineimage{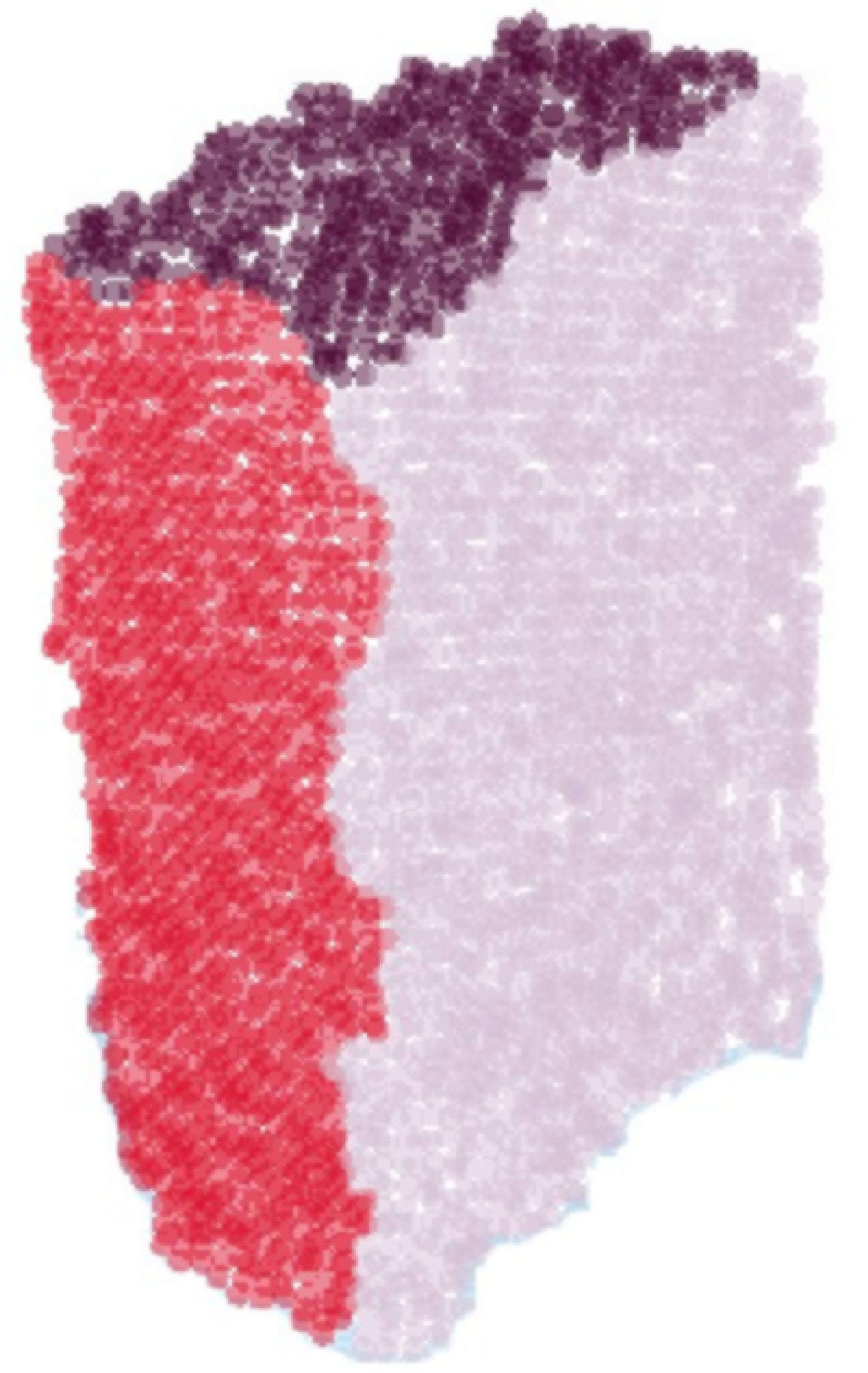} (consisting of three segments) through $\mathcal{H}$ is shown in \subref{fig:ch_illustration}.
		Feasible propagations, which have been previously encoded in the hierarchy during the training phase but which are not affected by the \emph{box}, are depicted as \protect \inlineimage{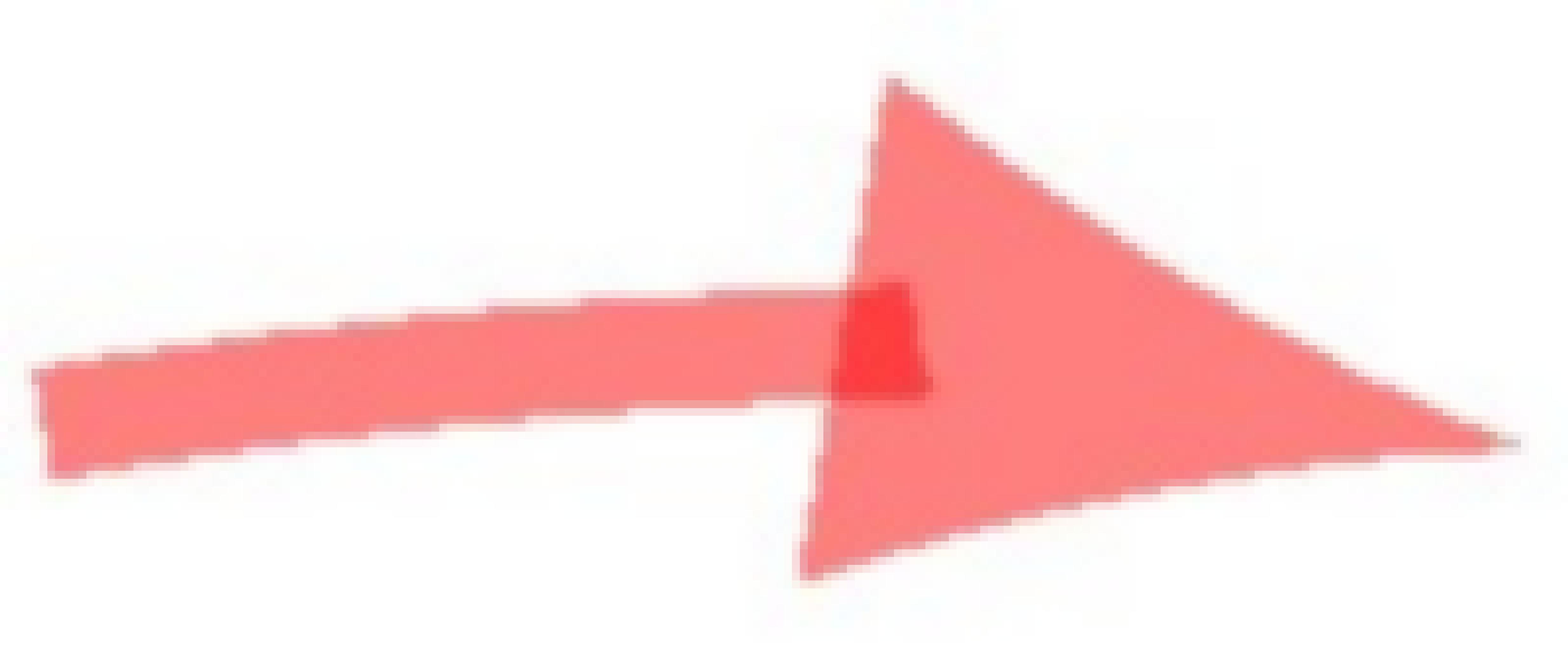}.
		Components of a \emph{motif level} are illustrated in \subref{fig:ch_legend}.  
		In \subref{fig:hch_illustration} the combined approach is illustrated:
		an example shape motif hierarchy ensemble $\mathcal{HE}$ based on three shape motif hierarchies $\{\mathcal{H}_1, \mathcal{H}_2 ,\mathcal{H}_3\}$ using respective description levels $\{d_1, d_2, d_3\}$ of $\mathcal{D}$ (see Fig.~\ref{fig:dictionary}).
	}
	\label{fig:dict_illustration_and_ch}
\end{figure}

The spatial topology of segments is analyzed in an unsupervised manner and encoded in a hierarchical representation, which we denote as \emph{Shape Motif Hierarchy}; an illustration of a hierarchy $\mathcal{H}$ is shown in Fig.~\ref{fig:ch_illustration}.
$\mathcal{H}$ is based on a graphical representation of visual word constellations which are denoted as \emph{motifs}; note that these constellations can only contain visual words of a specific description level. 
Therefore for a dictionary $\mathcal{D}\mathrm{=}\{d_1,d_2, ..., d_n\}$ which contains $n$ description levels, $n$ hierarchies are created that constitute an ensemble  $\mathcal{HE}\mathrm{=}\{\mathcal{H}_1,\mathcal{H}_2, ..., \mathcal{H}_n\}$, see Fig.~\ref{fig:hch_illustration}.

In the training phase for each hierarchy $\mathcal{H}$, object observations are encoded in a bottom-up manner, beginning with single object segments over groups of segments until a single constellation of segments represents the entire object. A sample propagation (\protect \inlineimage{sample_ch_prop_object}) of a box \inlineimage{sample_object_a} (consisting of three segments) through the hierarchy $\mathcal{H}$ is shown in Fig~\ref{fig:ch_illustration}.
Object segments are propagated through the hierarchy using the corresponding visual words associated to the segments.
Within the propagation process, newly observed visual word constellations (\emph{word motifs}) are integrated into the hierarchy as \emph{motif vertices} (see Fig.~\ref{fig:ch_legend}).
Each motif in the hierarchy is unique with respect to visual words, i.e., a newly observed word motif of an object leads to a creation of a \emph{motif vertex} if the motif does not exist in the hierarchy.
For further characterization of a motif vertex, 
a point cloud description is extracted of a propagated segment constellation and added as \emph{motif prototype} \samplePattern\ to the motif vertex \inlineimage{sample_clique_vertex_a} that corresponds to the motif of the propagated constellation (Fig.~\ref{fig:ch_legend}).
As a result, each motif vertex represents a \emph{shape motif} that can be exploited as \emph{building block} and that can constitute -- even unknown -- objects.
Further at motif level $l\mathrm{=}1$, an edge (\inlineimage{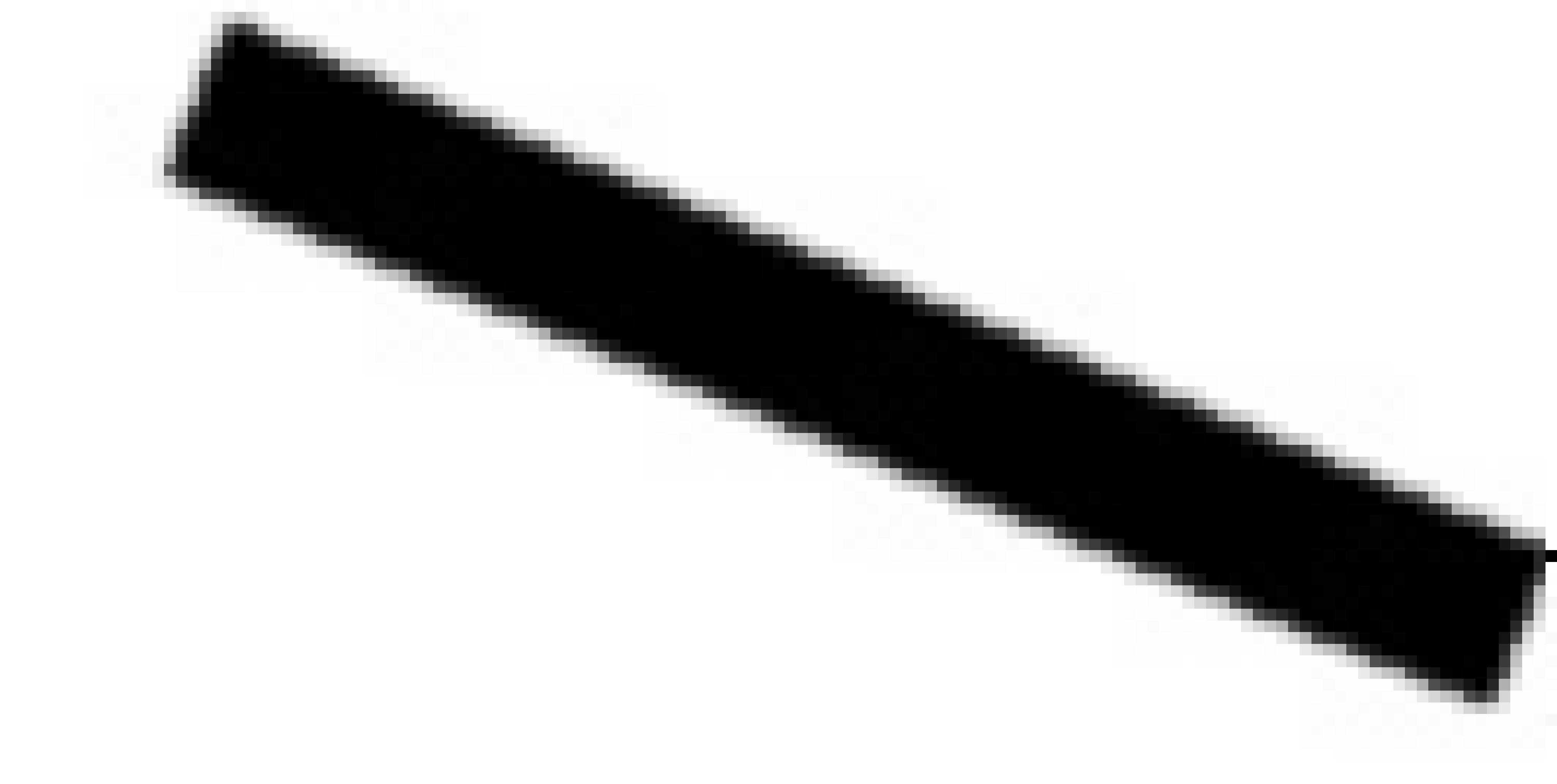}) between two motif vertices is created if the corresponding object segments are neighbors. For $l\mathrm{>}1$ an edge is created if two motif vertices contain a visual word that corresponds to the same segment of the propagated object.
In each propagation step from level $l$ to $l\mathrm{+}1$, the union of word motifs connected to an edge in level $l$ forms a vertex in $l\mathrm{+}1$ (\protect \inlineimage{sample_ch_prop_object}).
Consequently, upper levels can consist of fewer edges or vertices, i.e., a single motif vertex can encompass a word constellation that represents an entire object; see, e.g., the \emph{box} sample \inlineimage{sample_object_a} at motif level 4 in Fig.~\ref{fig:ch_illustration}.
In this manner, objects are decomposed in various motifs by the propagation through the hierarchy $\mathcal{H}$.

\begin{figure}[htbp]
	\small
	\centering
	\definecolor{visugraf_blue_1}{RGB}{64,64,160}
	\definecolor{visugraf_blue_2}{RGB}{64,94,255}
	\definecolor{visugraf_blue_3}{RGB}{64,223,255}
	\definecolor{visugraf_green_4}{RGB}{158,255,155}
	\definecolor{visugraf_yellow_5}{RGB}{255,237,64}
	\definecolor{visugraf_red_6}{RGB}{255,117,64} 	
	\includegraphics[width=0.99\linewidth]{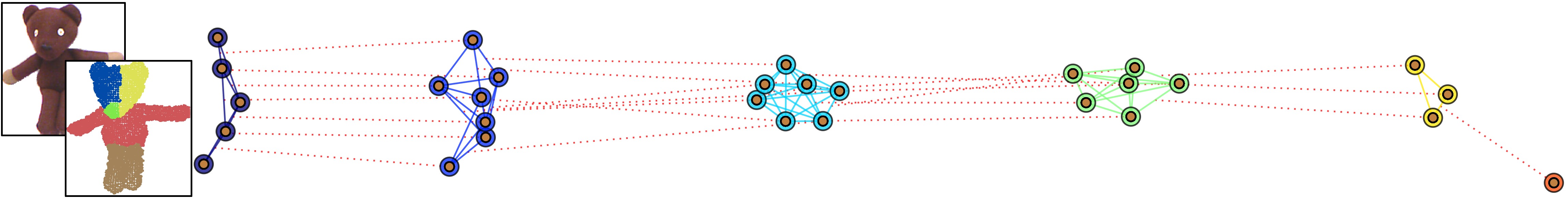}
	\put(-120,-20){\includegraphics[height=0.8cm]{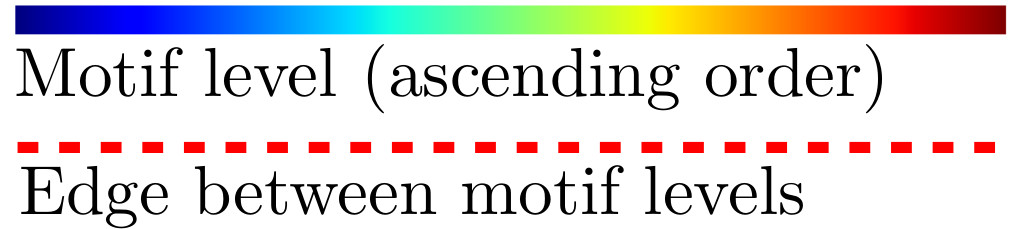}}
	\caption{Illustration of a segmented teddy sample propagated through $\mathcal{H}_4$ (see Fig.~\ref{fig:ch_illustration}) showing \emph{activated} motif vertices in each motif level (see Fig.~\ref{fig:ch_legend}), i.e. level \textcolor[RGB]{64,64,160}{\textbf{1}}, \textcolor[RGB]{64,94,255}{\textbf{2}}, \textcolor[RGB]{64,223,255}{\textbf{3}}, \textcolor[RGB]{158,255,155}{\textbf{4}}, \textcolor[RGB]{255,237,64}{\textbf{5}} and \textcolor[RGB]{255,117,64}{\textbf{6}} (the teddy segments in the point cloud are randomly colored).
	}
	\label{fig:simu:teddy_HE}
\end{figure}

As a result, the Shape Motif Hierarchy Ensemble $\mathcal{HE}\mathrm{=}\{\mathcal{H}_1,\mathcal{H}_2, ..., \mathcal{H}_n\}$ does not only take the structural appearance with respect to the variety of the segment constellations into account but also the symbolic appearance of constellations by using a specific dictionary description level for the respective hierarchy.
For illustration purposes, the propagation process of a segmented \emph{teddy bear} is shown in Fig.~\ref{fig:simu:teddy_HE} from a set of primitive motif vertices to more complex motifs vertices in which eventually a single motif represents the \emph{teddy bear}.

\subsection{Stimuli Generation} %
\label{sec:stimuli_generation}
In the training phase, object segment constellations represented by corresponding visual words are propagated through the hierarchy and are memorized as \emph{motif prototypes} within motif vertices that match visual word constellations of the object. %
Inspired by the \emph{Prototype Theory}~\cite{Rosch1973}, each motif vertex is formed by these prototypes, which are used to generate stimuli for unknown objects as described in the following: given a graph of segments $g^o$ of object $o$, the segments are annotated with the corresponding words and subsequently propagated through the hierarchy as in the training phase, see the \emph{box} example in Fig.~\ref{fig:ch_illustration} -- note that the hierarchy is not modified during the stimuli generation.
Through the propagation of segments, motif vertices are activated that correspond to the words of the propagated segments.
An activation of a vertex $v$ is represented by the Indicator function $\mathds{1}_v(g^o)$, which returns $1$ in case of a match, otherwise $0$ if no match is found.
For an activated $v$, a stimulus $\alpha(v,g^o)$ is computed based on point cloud descriptions of the memorized motif prototypes $T^v$ of $v$ and the respective description $q$ of object segments in $g^o$,
which activated $v$.
By applying Probabilistic Neural Networks~\cite{Huang:2004:APN:1011980.1011984}, the stimulus is computed with an adapted Gaussian kernel (bandwidth $\sigma\mathrm{=}0.025$) in which Jenson-Shannon divergence~($JSD$)~\cite{lin1991divergence} 
is used as distance measure, see Eq.~\ref{eq:stimuli}.
\begin{equation} \label{eq:stimuli}
\small
\alpha(v,g^o) = 
\begin{cases}
    \frac{1}{|T^v|} \cdot \sum^{|T^v|}_{i=1}e^{\tfrac{\mathrm{JSD}(t_i \in T^v,q)^2}{-2\sigma^2}},&\text{\footnotesize if $\mathds{1}_v(g^o)\mathrm{=}1$}\\
    0,              & \text{\footnotesize otherwise}
\end{cases}
\end{equation}
As a result for each propagated object, stimuli of motif vertices in $\mathcal{H}_i$ are accumulated and projected into vector form $\gamma^o_i\mathrm{=}[\alpha(v_1,g^o),  \alpha(v_2,g^o), ...]$.
Given $n$ description levels and correspondingly trained $n$ shape motif hierarchies that form the ensemble $\mathcal{HE}\mathrm{=}\{\mathcal{H}_1,\mathcal{H}_2, ..., \mathcal{H}_n\}$, the object graph $g^o$ is propagated through each motif hierarchy. 
Subsequently, a final stimuli vector $\leftidx{^*}\gamma^o\mathrm{=}[\gamma^o_1$, $\gamma^o_2$, $..., \gamma^o_n]$ is composed ($||$) of stimuli retrieved from $n$ motif hierarchies, see Fig.~\ref{fig:hch_illustration}.

\section{Descriptive Topology Analysis}
\label{sec:desc_topo_analysis}
Commonalities among shape appearances can vary from \emph{specific} to \emph{generic} shape facets: a concept generation process is hence used, which in a gradual manner detects commonalities ranging from individual to common facets, i.e., very specific to often re-occurring facets.
Persistent Homology (PH) provide the computational model that allows to gradually reveal topologically persistent patterns in  generated stimuli $\leftidx{^*}\gamma$, which are interpreted as commonalities and eventually as shape concepts.

\subsection{Persistence Homology and Filtration}
\label{sec:filt_PH}
We briefly introduce terms from algebraic topology which are related to our shape concept learning approach.
Comprehensive literature can be found in \cite{carlsson2014,7487710,Edelsbrunner2002,Zomorodian:2005:CPH:1044838.1044846,Zhu:2013:PHI:2540128.2540408}.

\subsubsection{Simplices and Complexes}
Given a continuous topological space $\mathcal{X}\mathrm{=}\{ x_0, x_1, ..., x_m| x_i\in \mathcal{R}^n, 0 \leq i \leq m\} $ with $m$ $n$-dimensional data points.
A \emph{simplex} $\pi$ is a $d$-dimensional polytope, which is a graph consisting of a convex hull of $d\mathrm{+}1$ affine independent vertices where each vertex is a point in $\mathcal{X}$.
A composition of \emph{simplices} is denoted as \emph{simplicial complex} $K\mathrm{=}\{\pi_0, \pi_1, \pi_2, ...\}$.
This composition is a union of vertices, edges, triangles or other higher dimensional polytopes.
\subsubsection{Vietoris-Rips Complex}
We focus on \emph{vietoris-rips complexes} in which a complex $K^{vr}_i$ is extracted from a subspace $\mathcal{X}_i \mathrm{\subseteq} \mathcal{X}$ with a given scale parameter $\epsilon \mathrm{>} 0$.
$K^{vr}_i$ consists of vertices that are only connected if the distances between the vertices is lower than the given parameter $\epsilon$.
The vietoris-rips complex $K^{vr}_i$ can also be denoted as $\epsilon$\emph{-complex}, where $\epsilon$ is also denoted as radius or distance threshold.

\subsubsection{Homology Groups}
Homology is a concept in algebraic topology, which allows to reveal specific characteristics or features in $\mathcal{X}$.
Characteristics are organized therein into homology groups $\mathcal{HG}\mathrm{=}\{H_0(\mathcal{X}), H_1(\mathcal{X})$, $H_2(\mathcal{X}), ...\}$.
Often, the first three homology groups are analyzed: in the context of geometry $H_0(\mathcal{X})$ is related to \emph{connected components} or \emph{clusters} of vertices. $H_1(\mathcal{X})$ is related to the complexes in form of \emph{loops} or \emph{holes} and $H_2(\mathcal{X})$ is related to \emph{voids} which represent fully connected complexes.
Here, we focus on $H_0(\mathcal{X})$ since it complies with our goal to extract topological groups from stimuli vectors (see Sec.~\ref{sec:stimuli_generation}), which can represent concepts.
\subsubsection{Topological Space Filtration and Persistent Homology}
\label{sec:topo_filration}
The filtration of the topological space $\mathcal{X}$ is initiated by a subsequently nested application of a set of radii $\mathcal{E}\mathrm{=}\{\epsilon_0, \epsilon_1, ..., \epsilon_j\}$ where $\epsilon_{i-1} \mathrm{<} \epsilon_i \mathrm{<} \epsilon_{i+1}$.
For $H_0(\mathcal{X})$, each point $x_i \in \mathcal{X}$ is represented at the beginning of the filtration process by a 0-simplex $\pi_i \in$ \emph{vietoris-rips complexes} $K^{vr}_0$. %
These simplices are so to say \emph{born} at radius $0$.
Note that the $K^{vr}_i$ is extracted using radius $\epsilon_i$.
While the filtration progresses, the vietoris-rips complex grows since the radius increases, which can cause fusions of simplices that form a larger simplex: a \emph{union} is performed between simplices while one simplex enlarges and sustains by annexing the other that \emph{dies}.
Eventually, a complex $K^{vr}$ is filtered that contains a single high dimensional simplex -- see Eq.~\ref{eq:filtration}.
\begin{equation}
\small
\emptyset \subseteq K^{vr}_0 \subseteq K^{vr}_1 \subseteq \dotso \subseteq K^{vr}_j = K^{vr}
\label{eq:filtration}
\end{equation}
Persistent Homology provides a way to analyze and track birth and death of simplices (also known as \emph{homology classes}) along the filtration process: $H_0(K^{vr}_i) \mathrm{\rightarrow}  H_0(K^{vr}_{i+1})$.
The according results can be represented in \emph{persistence} or \emph{barcode diagrams} (Fig.~\ref{fig:barcode}).
While considering the gradual evolution of vietoris-complex $K^{vr}_i$, the extraction of homology classes (birth and death) is inherently robust to deformation due to the topological organization of the data in a graphical manner.

\subsection{Shape Concept Extraction}
\label{sec:concept_extraction}
In the following, the shape concept extraction process is described -- from topological space and concept generation to concept inference.
\subsubsection{Topological Space Generation}
\label{sec:topo_space_gen}
Given a set of raw stimuli vector responses (Sec.~\ref{sec:stimuli_generation}), the responses are initially used to create a topological space in a graphical manner.
Therein, a stimuli vector  $\leftidx{^*}\gamma$ can be interpreted as an independent point in the space, in which a distance metric can be used to measure the similarity to other stimuli vectors; these vectors serve as anchor points in a space of an unknown topology.
The goal is to interrelate these vectors in order to discover topological relationships among the anchor points.
We make use of a graphical representation, in which each anchor point represents a vertex.
Initially a complete graph is created, where each edge between vertices is augmented with the corresponding distance; distances are measured by the Jenson Shannon divergence (JSD). %

\begin{figure}[htbp]
   \small
  \centering
  \includegraphics[width=0.85\linewidth]{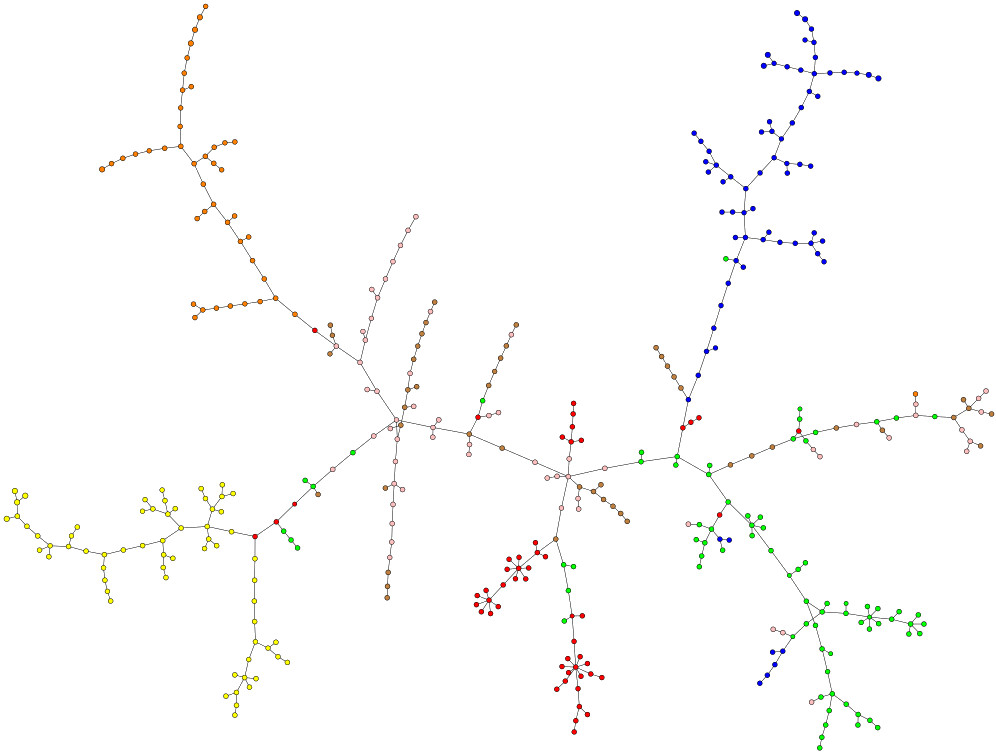}
  \includegraphics[trim=0.3cm 0.1cm 0.1cm 0.1cm,width=0.1385\linewidth]{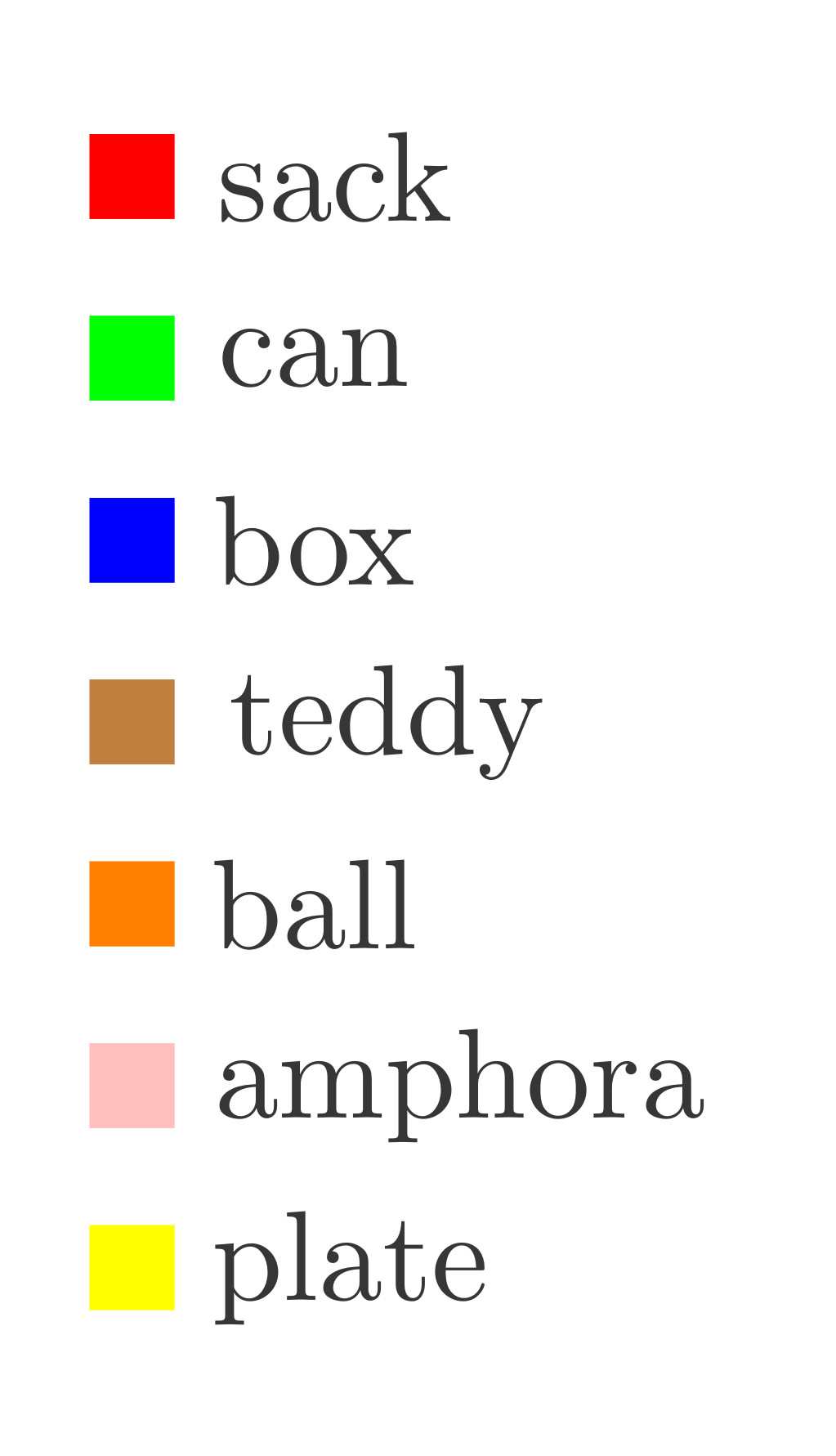}
  \caption{The minimum spanning tree, which spans the topological space $\mathcal{X}$ of stimuli vectors extracted from instances of the \emph{Object Shape Category Dataset} (OSCD), see Sec.~\ref{sec:experiment}. Note that each vertex represents a sample object of the dataset. Vertices are colored only for illustration purposes by their corresponding category label of the dataset, which is not used in our unsupervised learning phase.}
  \label{fig:mst}
 \end{figure}

To minimize the search space and to initiate the construction of the topological space $\mathcal{X}$, the Minimum Spanning Tree~\cite{kruskal_mst} is extracted using the respective JSD distances.
Subsequently, a substantial amount of edges perishes and a minimum number of edges remain, which reveal the structural and topological organization of the stimuli vectors.
Fig.~\ref{fig:mst} shows an example based on the object instances from the \emph{Object Shape Category Dataset} (Sec.~\ref{sec:experiment}) that consists of seven shape categories  (\emph{sack}, \emph{can}, \emph{box}, \emph{teddy}, \emph{ball}, \emph{amphora}, \emph{plate}). 

From this point on, we focus on the \emph{topological similarity} among stimuli in form of the \emph{geodesic} distance within $\mathcal{X}$. %
Therefore each edge is uniformly weighted by assigning a distance of $1$. 
Due to the inherent sparsity of edges in $\mathcal{X}$, Johnsons all-pair-shortest path algorithm allows to efficiently generate a distance map which is used to infer a heat for each vertex $x \in \mathcal{X}$.
A vertex heat $h^{\simpcir}(x)$ is inferred by the mean geodesic distances $d_{geo}(\cdot)$ to all other vertices in $\mathcal{X}$ whereas the edge heat $h^{\ue}(e_{j,k})$ is determined by the mean heat of the connected vertices $x_j$ and $x_k$ as shown in Eq.~\ref{eq:vertex_heat}.

\begin{equation}
\small
h^{\simpcir}(x)\mathrm{=}\frac{\sum_{i=0}^{|\mathcal{X}|} d_{\text{geo}}(x, x_i\in\mathcal{X})}{|\mathcal{X}|},\ \ h^{\ue}(e_{j,k})\mathrm{=}\frac{h^{\simpcir}(x_j)\text{+}h^{\simpcir}(x_k)}{2}
\label{eq:vertex_heat}
\end{equation}
Henceforth, we use edge heats as edge distances between respective vertices. 
By scaling the heat in $\mathcal{X}$ to the interval $[0,1]$ and inverting the heat,
vertices located at leaf regions of $\mathcal{X}$ come closer to each other whereas vertices in the inner region move farther away from each other.
Furthermore, two observations can be made: a) the heat of exteriorly located edges is lower than the interiorly located ones; b) vertices which are interiorly located reflect more heterogeneity with respect to their neighbors, compared to vertices which are exteriorly located in $\mathcal{X}$.

\subsubsection{Topological Filtration}
\label{sec:shape_filtration}
Given the topological space $\mathcal{X}$, the filtration is applied over a range of radii $\mathcal{E}\mathrm{=}\{\epsilon_0, \epsilon_1, ..., \epsilon_j\}$.
The step size $\epsilon_i \mathrm{\rightarrow} \epsilon_{i+1}$ is determined by the minimum edge distance in $\mathcal{X}$ that also initializes the filtration at $\epsilon_0$.
The filtration is completed when the maximum edge distance in $\mathcal{X}$ is reached at $\epsilon_j$. 
In practice, the number of steps $|\mathcal{E}|$ can reach a computationally intractable number.
An upper bound limit for $|\mathcal{E}|$ can be applied by increasing the step size until the upper bound is met.
Consequently, the filtration is initialized with 0-simplices where each simplex represents a stimuli vector, i.e., a vertex of the topological space $\mathcal{X}$.
This filtration is performed on $\mathcal{X}$ as described in Sec.~\ref{sec:topo_filration}; note that the equidistant filtration steps from $\epsilon_0$ to $\epsilon_j$ are often denoted as time.

\begin{figure}[htbp]
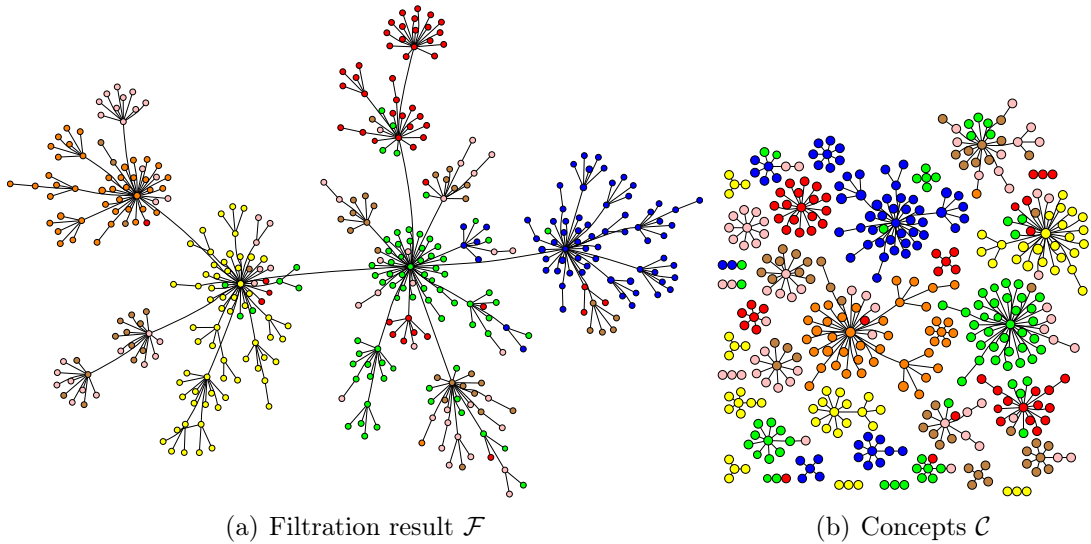

   \small
   \centering
	\subfigure[Filtration result $\mathcal{F}$]{\label{fig:association_graph}\includegraphics[width=0.64\linewidth]{association_graph_no_label}} 
	\subfigure[Concepts $\mathcal{C}$]{\label{fig:ph_concept_extraction}\includegraphics[width=0.345\linewidth]{concept_association_graph_no_label}}
  \caption{\subref{fig:association_graph} A filtration graph $\mathcal{F}$ showing annexations over time according to the given graph $\mathcal{X}$ (see Fig.~\ref{fig:mst}). For illustrations purposes, each vertex is colored with the corresponding label as shown in Fig.~\ref{fig:mst}. \subref{fig:ph_concept_extraction} Connected components extracted from $\mathcal{F}$ \subref{fig:association_graph} that represent concepts $\mathcal{C}$ ($|\mathcal{C}|\mathrm{=}36$). For illustration purposes, each vertex (concept prototype) is colored with the corresponding label as in Fig.~\ref{fig:mst}.}
  \label{fig:association_graph_ph_concept_extraction}
 \end{figure}

Persistent Homology allows to track the birth and death of simplices in $K^{vr}$ of $\mathcal{X}$ during the filtration.
Due to the nature of evolving simplices complexes (see Eq.~\ref{eq:filtration}) in each time step, the complex changes its appearance after annexations of simplices complexes of previous time steps.
These changes during the filtration are encoded in graph $\mathcal{F}$, which is shown in Fig.~\ref{fig:association_graph}.

An edge represents an annexation during the filtration process of a simplices complex to another complex -- beginning with 0-simplices representing leaves in $\mathcal{F}$. 
Each edge is augmented with the annexation time.
So, outer simplices lived shorter since they have been annexed earlier in time compared to inner ones.
	As a result, $\mathcal{F}$ represents the filtration progression of $\mathcal{X}$.
\subsubsection{Extraction of Persistent Shape Concepts}
\label{sec:shape_concept_extraction}
The lifetime of simplices can be interpreted as a feature indicator in $\mathcal{X}$, i.e., \emph{persistent} or long living simplices tend to represent a significant feature, i.e., a 
shape property that is prominent for an object or even object category.
At the same time, short living simplices can be interpreted as being insignificant.
The goal is hence to detect persistent simplices.
In order to ease the persistence analysis, the filtration time range is scaled within the interval $[0,1]$, i.e., from $0$ (start of filtration $\mathrm{=} \ \epsilon_0$) to $1$ (end of filtration $\mathrm{=} \ \epsilon_j$).
In the filtration process, trivial homology classes are obtained at time $0$ where 0-simplices exist and at time $1$ where a single simplex consists of all simplices in $\mathcal{X}$.
We are interested of finding persistent groups between these extrema.
A \emph{group} is a connected component of vertices, i.e., a $d$-simplex ($d\mathrm{>}0$).
Due to the gradual filtration, each group consists of topologically similar vertices. 
Therefore, the groups can constitute shape concepts, where each vertex within a group is a representative \emph{concept prototype}.

Given the entire time spectrum $[0,1]$, Persistent Homology allows to access any state of detected concepts $\mathcal{C}$ in $\mathcal{X}$ at an arbitrary time in the spectrum; note that the filtration starts with $|\mathcal{C}|\mathrm{=}|\mathcal{X}|$ and ends with $|\mathcal{C}|\mathrm{=}1$.
Consequently, a distinctive time can be determined.
An optimal time varies according to the topology that is reflected by the given stimuli vector.
Consider an optimal time when the global maximum of annexations (see Sec.~\ref{sec:eval:filtration}) is reached, and subsequently edges in $\mathcal{F}$ that are augmented with an older time than the optimal time are removed. %
This optimal time leads to a set of connected components in $\mathcal{F}$ that can reflect useful shape concepts as illustrated in Fig.~\ref{fig:ph_concept_extraction}.
Note that edges which are created at later time connect more heterogeneous groups and subsequently represent more, and possibly too generic concepts, in contrast to more specific concepts which emerge when edges are created at earlier time.

\subsection{Shape Concept Inference}
\label{sec:shape_concept_inference}
Given a stimuli vector $\leftidx{^*}\gamma^o$ that is extracted from an unknown object $o$,  a response is retrieved based on similarity to previously learned shape concepts (see Fig.~\ref{fig:ph_concept_extraction}).
Each concept $c\in\mathcal{C}$ consists of a set of \emph{concept prototypes} $P^c\mathrm{=}\{p_1,$ $p_2, ...\}$, which are used to derive the correspondence of unknown objects to concepts.
In the spirit of \emph{Prototype Theory} \cite{Rosch1973}, unknown instances are classified based on the similarity to known instances, which are associated to the previously learned shape concepts.
To demonstrate the discrimination capability of our shape representation, the similarity $\phi^c(\cdot)$ to a concept $c$ is determined by a (basic) mean similarity among  $\leftidx{^*}\gamma^o$ and prototypes $P^c$ of concept $c$ (see Eq.~\ref{eq:concept_response}); as distance measure the \emph{Mahalanobis distance}
$d_{\text{mah}}(\cdot)$ is used.
\begin{equation}
\centering
\small
\phi^c(\leftidx{^*}\gamma^o) = \frac{  \sum^{|P^c|}_{i=1} d_{\text{mah}}( \leftidx{^*}\gamma^o, p_i \in P^c )}{|P^c|}
\label{eq:concept_response}
\end{equation}

\section{Machine-Centric Concept Generation Through Mental Simulation}
\label{sec:machine_centric_concepts}

An interesting question is how the training data is generated to learn the shape concepts. 
One option is to use datasets of real-world objects.
In contrary, we propose the use of mental simulation to generate \emph{abstract artificial} objects for shape concept learning purposes.
This approach generates concepts in a machine-centric manner, i.e., concepts are learned in an  unsupervised fashion in two respects: a) \emph{label-agnostic} (no label information given by supervision is used) and additionally b) \emph{instance-agnostic} (no real-world instances preselected by human supervision are used for training).
As will be shown in the experiments in Sec.\ref{sec:simu:experiment}, the shape concepts learned in this way generalize well when applied to objects from real-world datasets.

\begin{figure}[tb]
	\def\scalesample{2.7cm}
	\centering
	\subfigure[]{\label{fig:demo:unsup_sim:sample1}\includegraphics[height=\scalesample]{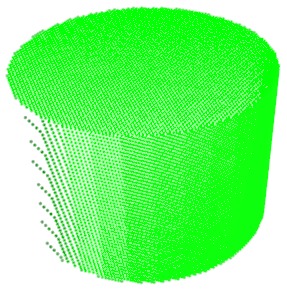}}
	\subfigure[]{\label{fig:demo:unsup_sim:sample2}\includegraphics[height=\scalesample]{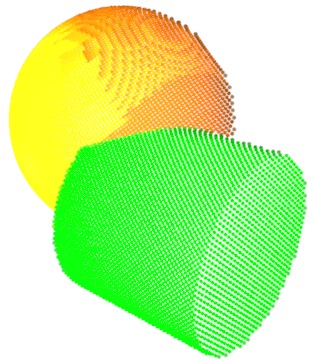}}
	\subfigure[]{\label{fig:demo:unsup_sim:sample3}\includegraphics[height=\scalesample]{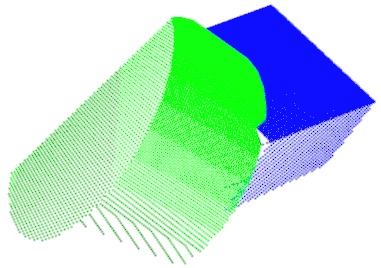}}	
	\subfigure[]{\label{fig:demo:unsup_sim:sample4}\includegraphics[height=\scalesample]{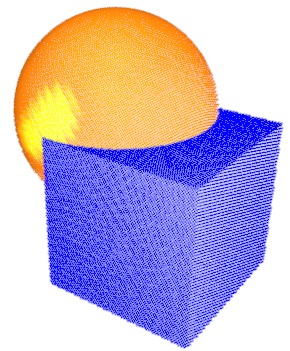}}	
	\subfigure[]{\label{fig:demo:unsup_sim:sample5}\includegraphics[height=\scalesample]{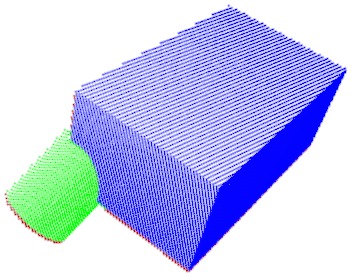}}	
	\subfigure[]{\label{fig:demo:unsup_sim:sample6}\includegraphics[height=\scalesample]{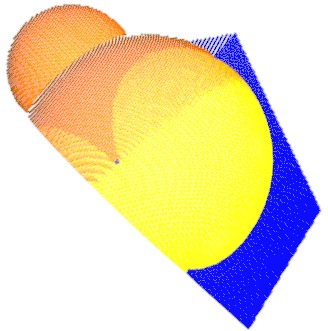}} 	
	\subfigure[]{\label{fig:demo:unsup_sim:sample7}\includegraphics[height=\scalesample]{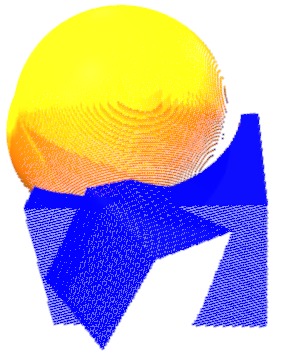}}
	\subfigure[]{\label{fig:demo:unsup_sim:sample8}\includegraphics[height=\scalesample]{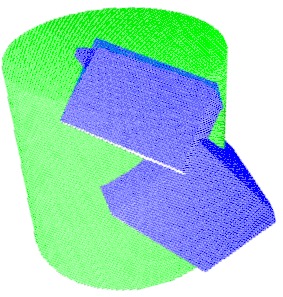}}
	\subfigure[]{\label{fig:demo:unsup_sim:sample9}\includegraphics[height=\scalesample]{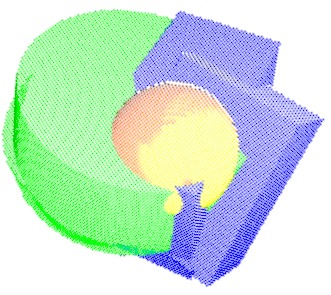}}
	\caption{Examples of randomly generated abstract objects. Objects can encompass up to five primitive-shaped prototypes. Only for illustration purposes, the primitive-shaped prototypes of each object are distinctively colored: \textcolor[RGB]{0,0,240}{\textbf{box}}, \textcolor[RGB]{0,240,0}{\textbf{can}} and \textcolor[RGB]{240,240,0}{\textbf{sphere}}. }
	\label{fig:demo:unsup_sim:samples}
\end{figure}

The core idea for the mental simulation is described in the following. 
We start with primitive-shaped building-blocks or prototypes, namely \emph{box}, \emph{sphere}, and \emph{cylinder}. 
Multiple prototypes can be randomly combined to a prototype composition which forms an abstract object. 
We denote the number of introduced prototypes of an abstract object as the \emph{prototype order}. 
The Gazebo simulation environment~\cite{Koenig2004} is then used to generate these artificial abstract objects in simulation and to capture samples of the generated objects with a virtual sensor in simulation.
Fig.~\ref{fig:demo:unsup_sim:samples} shows samples of artificial abstract objects of different prototype orders, captured in simulation.

\begin{algorithm}[htbp]
	\footnotesize
	\caption{\small Artificial Sample Generation}
	\label{alg:art_sample_generation}   
	\begin{algorithmic}[1]
		\footnotesize
		\floatname{algorithm}{Procedure}
		\renewcommand{\algorithmicrequire}{\textbf{Input:}}
		\renewcommand{\algorithmicensure}{\textbf{Output:}}
		\REQUIRE prototype order $n$, empty sample $s$
		\STATE $i$ $\gets$ 0
		\WHILE{$i < n$ }
		\STATE $p$ $\gets$ get\_random\_prototype(\{\emph{box}, \emph{cylinder}, \emph{sphere}\})
		\STATE $p$ $\gets$ set\_random\_dimensions(\{\emph{length}, \emph{width}, \emph{height}, \emph{radius}\})
		\IF{$i > 0$}
		\STATE $p$ $\gets$ set\_random\_pose(\{position, orientation\}) \\ \hspace{16pt} so that $p$ intersects with $s$
		\ENDIF
		\STATE $s$ $\Longleftarrow$ $p$ (introduce prototype $p$ to sample $s$)
		\STATE$i$ $\gets$ $i\mathrm{+}1$
		\ENDWHILE
		\ENSURE sample $s$ representing a composition of prototypes.
	\end{algorithmic}
\end{algorithm}

Each prototype of an artificial object sample is not only randomly generated with respect to its type (\emph{box}, \emph{sphere}, \emph{cylinder}) but also with respect to its spatial dimensions (e.g., \emph{length}, \emph{width}, \emph{height}, \emph{radius}). Each prototype has to overlap with at least an other prototype in order to form a connected structure, which is considered as a valid object (see Alg.~\ref{alg:art_sample_generation}).
Using this random approach, these object samples are obviously generated without any human bias.

\section{Experiments: Label-Agnostic Learning} 
\label{sec:experiment}

\begin{figure}[tb]
	\small
	\centering
	\def \comImgHeight {0.125}
	\subfigure[]{\label{fig:trainsack}\includegraphics[height=\comImgHeight\textwidth]{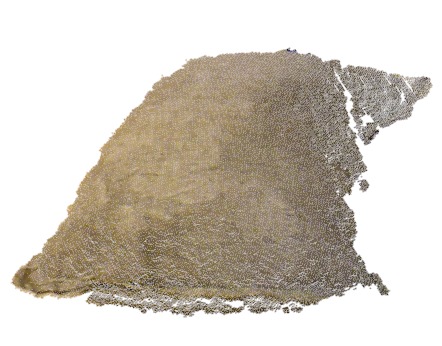}}
	\subfigure[]{\label{fig:trainbarrel}\includegraphics[height=\comImgHeight\textwidth]{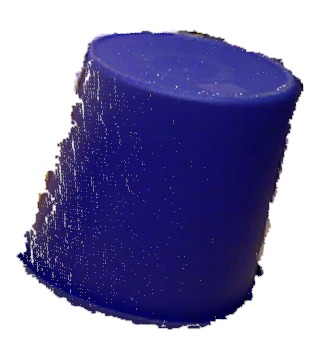}}
	\subfigure[]{\label{fig:trainparcel}\includegraphics[height=\comImgHeight\textwidth]{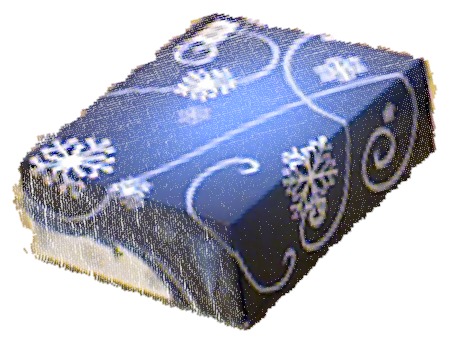}}
	\subfigure[]{\label{fig:trainteddy}\includegraphics[height=\comImgHeight\textwidth]{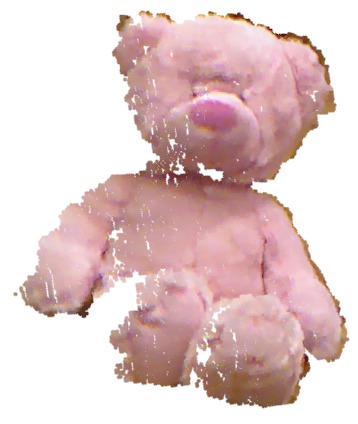}}
	\subfigure[]{\label{fig:trainball}\includegraphics[height=\comImgHeight\textwidth]{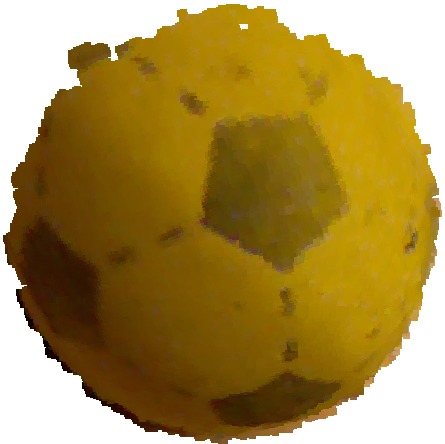}}
	\subfigure[]{\label{fig:trainamphora}\includegraphics[height=\comImgHeight\textwidth]{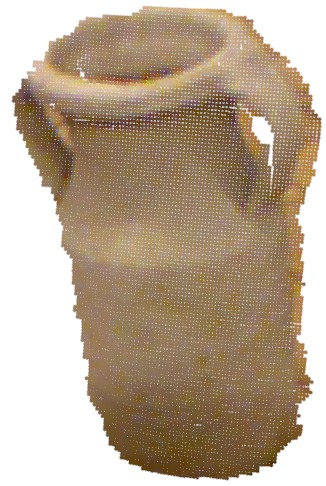}}
	\subfigure[]{\label{fig:trainplate}\includegraphics[height=\comImgHeight\textwidth]{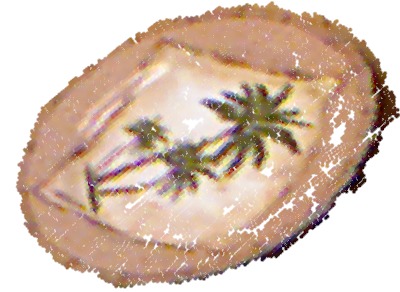}}
	\caption{%
	Examples of 2.5D scans from the OSCD dataset: \emph{sack}~\subref{fig:trainsack}, \emph{can}~\subref{fig:trainbarrel}, \emph{box}~\subref{fig:trainparcel}, \emph{teddy}~\subref{fig:trainteddy}, \emph{ball}~\subref{fig:trainball}, \emph{amphora}~\subref{fig:trainamphora} and \emph{plate}~\subref{fig:trainplate}.}
	\label{fig:ph_samples}
\end{figure}

The experimental evaluation is two-fold. This section deals with the performance of the label-agnostic concept generation. This means that the shape concepts are learned in an unsupervised manner; semantic object labels generated by humans are only used to evaluate how reasonable the generated concepts are. Among others, it will be shown that concepts learned on one real-world dataset also generalize well to other real-world datasets consisting of different objects. 
In the following Sec.~\ref{sec:simu:experiment}, the focus is on the evaluation of machine-centric learning of the shape concepts, i.e., not real-world data but abstract artificial objects from mental simulation are used for training, which also leads to concepts that also perform well on the real-world datasets.

The \emph{Object Shape Category Dataset (OSCD)}\footnote{http://www.robotics.jacobs-university.de/datasets/2017-object-shape-category-dataset-v01/index.php}~\cite{mueller_birk_iros2018} is used for the first part of the evaluation. It consists of 
468 RGBD scans of real-world objects from 7 categories. A few examples are shown in Fig.~\ref{fig:ph_samples}).

\subsection{Topological Filtration}
\label{sec:eval:filtration}

\begin{figure}[htbp]
  \small
  \centering  
   \subfigure[Barcode  (H$_0$)]{\label{fig:barcode}\includegraphics[height=0.41\linewidth]{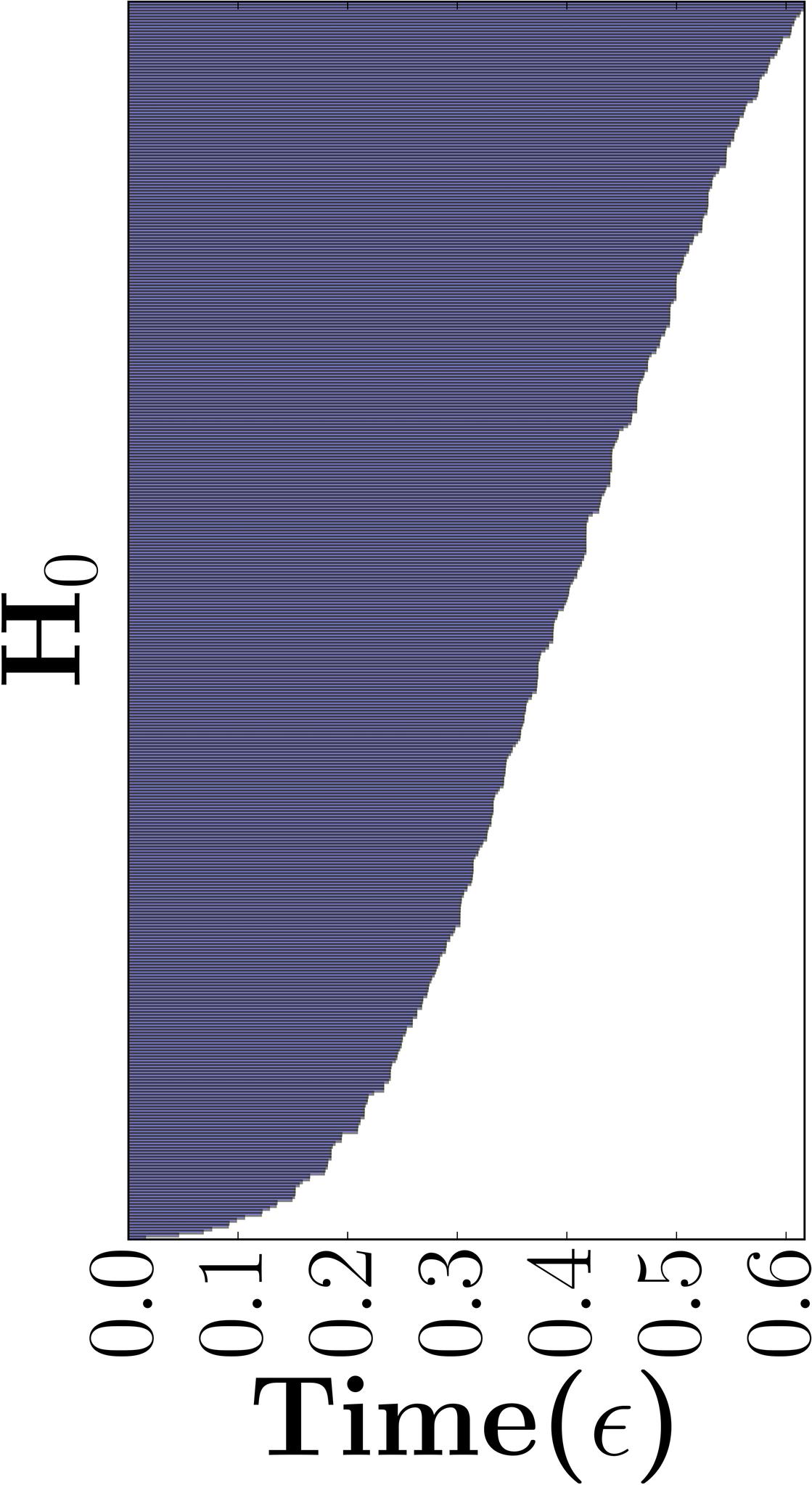}}\hfill
   \subfigure[Annexation]{\label{fig:annexation}\includegraphics[height=0.41\linewidth]{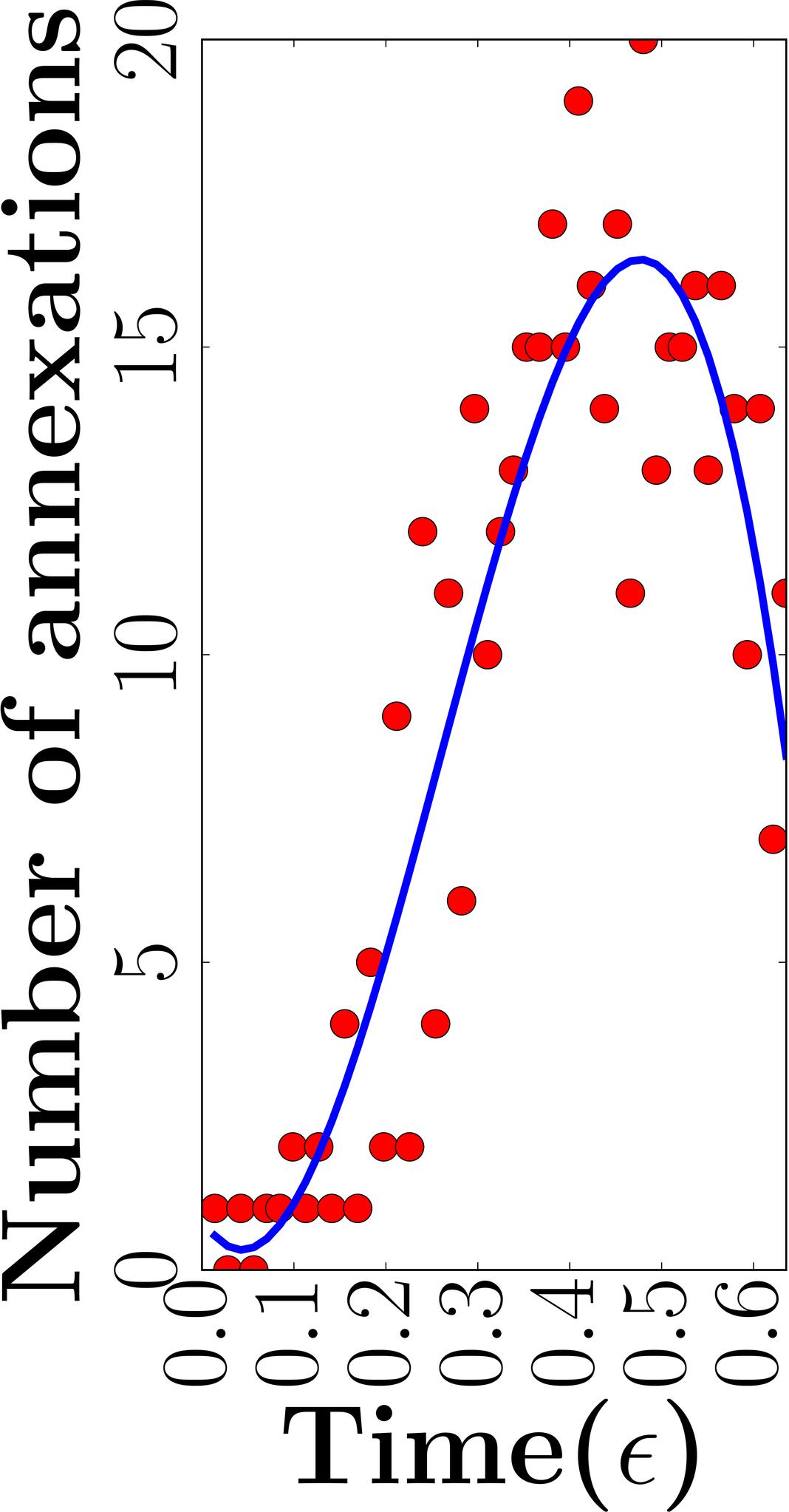}}\hfill
   \subfigure[Ranked concepts from Fig.~\ref{fig:ph_concept_extraction}.]{\label{fig:purity_proportion_result}\includegraphics[height=0.405\linewidth]{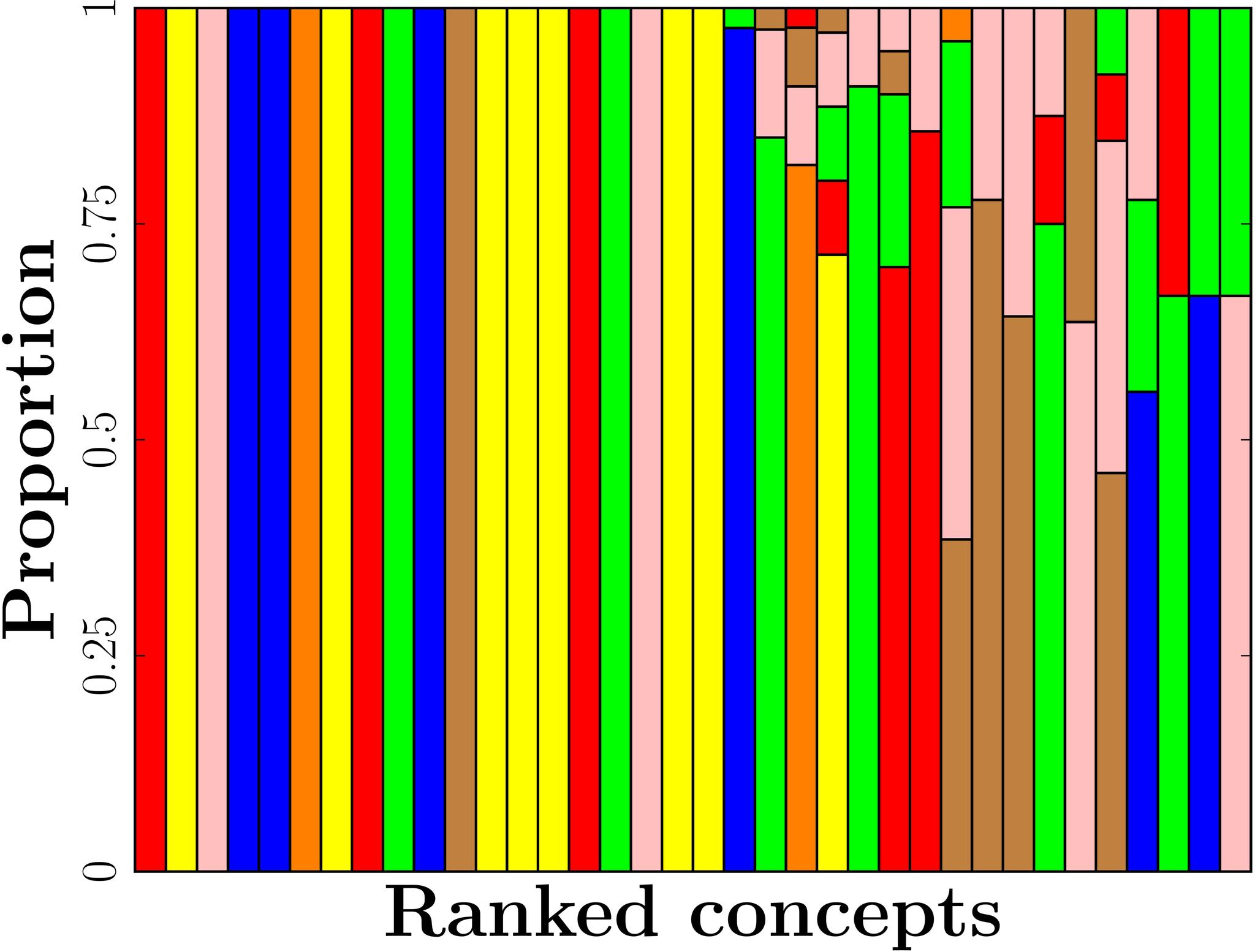}}
   \caption{\subref{fig:barcode} Barcode of the homology group 0. \subref{fig:annexation} The number of annexations among Homology classes. \subref{fig:purity_proportion_result} The proportional distribution of prototypes per concept. For visualization purpose, each proportion within a bar is colored with the corresponding label according to Fig.~\ref{fig:mst} and sorted in ascending order by $rs(\cdot)$, see Eq.~\ref{eq:rank_score}.}
   \label{fig:concept_generation_result}
\end{figure}

In the training phase, each training sample scan (OSCD dataset) is propagated through  $\mathcal{HE}\mathrm{=}\{\mathcal{H}_1, \mathcal{H}_2, ..., \mathcal{H}_n\}$, omitting any label-related information, i.e., each scan is applied in an unsupervised manner to the $\mathcal{HE}$; in our evaluation $n\mathrm{=}4$ has been heuristically selected -- a smaller $n$ may not allow $\mathcal{HE}$ to sufficiently discriminate the observed range of object shape variety.  
Afterwards, extracted stimuli vectors are fed to the filtration process (see Sec.~\ref{sec:desc_topo_analysis}).
Fig.~\ref{fig:association_graph} illustrates the filtration result of the stimuli vectors; the visualization does not reflect metric differences, it visualizes topological similarities among samples.
Already at this stage, topological similarity can be observed with respect to the category labels of the objects. 
Note that the category labels are only associated to the prototypes for visualization purposes - as mentioned, they were not used in the training.
In Fig.~\ref{fig:barcode}, the barcode is shown of the homology group 0.

At time $\epsilon_0$ all \emph{concept prototypes} -- depicted as bars -- are born. 
While the filtration progresses, more and more prototypes form larger homology classes that lead to the death (end of a bar) of prototypes, which have been annexed.
As a result, only a single simplex at time $\epsilon_j$ survives the filtration (see Sec.~\ref{sec:topo_filration}). 
Moreover, Fig.~\ref{fig:annexation} shows only the number of annexation of homology classes over time.
It can be observed that the filtration reaches a global maximum of annexations at $\epsilon_{max}\mathrm{=}0.48$, i.e., the annexation of classes decreases even though $\epsilon$ reaches its maximum value.
It can be interpreted that the extracted homology classes after $\epsilon_{max}\mathrm{=}0.48$ are already discriminative by their persistence.

\subsection{Unsupervised Concept Selection}
\label{sec:exp:unsup_concept_selection}
The gradual filtration process as described in Sec.~\ref{sec:filt_PH} allows to analyze the topological space at any filtration step. 
Each filtration step offers insights about the topology and the relation among concept prototypes. 
Note that the choice of a specific number of concepts and concept size depends on the objective of the application scenario. 

Using $\epsilon_{max}$ as indicator to stop the filtration process and to subsequently select the existing homology classes at time $\epsilon_{max}$ as concepts, we receive in total $36$ concepts $\mathcal{C}$ (see Fig.~\ref{fig:ph_concept_extraction}) with a minimum concept size of $2$.
To assess the quality of the extracted concepts we can make use of the human-annotated category labels, which are associated to the prototypes (see Fig.~\ref{fig:ph_concept_extraction}).
Therein, the correlation between the concepts and the labels given a priori by a human can be interpreted as a quality measure for the concepts learned in an unsupervised manner.
The amount of this correlation or \emph{purity} $pu(\cdot)$ can be defined as the largest proportion in the distribution of prototypes of a category label, see Eq.~\ref{eq:purity}, where concept $c\in\mathcal{C}$ consists of a set of concept prototypes $P^c\mathrm{=}\{p_1, p_2, ...\}$ which are accordingly attributed with labels $Y^c\mathrm{=}\{y_1, y_2, ...\}$, i.e., $y_i\mathrm{=} \mathrm{retrieve\_label}(p_i)$, given the set of category labels $\mathcal{Y}$ of the dataset where $y_i\in\mathcal{Y}$.
\begin{equation}
\small
\small
pu(c) = \argmax_{y\in\mathcal{Y}} \frac{\sum^{|P^c|}_{i=1}\mathds{1}_y(y_i\in Y^c)}{\big|P^c\big|}
\label{eq:purity}
\end{equation}

Given the concepts inferred by $\epsilon_{max}$ as described in Sec.~\ref{sec:shape_concept_extraction} and illustrated in Fig.~\ref{fig:ph_concept_extraction}, it can be observed that connected components of different sizes are extracted, which is caused by the shape heterogeneity of the prototypes in $\mathcal{X}$. 
A large portion of the concepts is pure (see Eq.~\ref{eq:purity}), i.e., there is a perfect correlation and only prototypes of a specific category $y\in\mathcal{Y}$ are assigned to a concept $c\in\mathcal{C}$.
In Fig.~\ref{fig:purity_proportion_result}, the resulting distribution of prototypes within a concept is illustrated.
Concepts are sorted in ascending order by the \emph{rank score} $rs(c)$, which computes the concept purity $pu(c)$ with respect to the concept size $\big|P^c\big|$, see Eq.~\ref{eq:rank_score}. 
\begin{equation}
\small
rs(c) = \frac{\big|P^c\big|}{1 - pu(c) + \varepsilon},\text{\footnotesize where } \varepsilon \mbox{\ \footnotesize is a small constant\ } (0\mathrm{<} \varepsilon \mathrm{\ll} 1)
\label{eq:rank_score}
\end{equation}
While $57\%$ of the concepts are pure, other concepts show a lower purity, i.e., samples of different categories are assigned to a particular concept. However, these categories show shape similarities like \emph{sack} and \emph{can} or \emph{plate} and \emph{box}. Furthermore, the mean concept purity is $86.2\%$.

Given the $36$ concepts, responses are extracted for each sample of the dataset, i.e., each sample object $o$ is represented by $\rho^o\mathrm{=}\{\phi^{1}( \leftidx{^*}\gamma^o), \phi^{2}(\leftidx{^*}\gamma^o),...\}$ ($|\rho^o|\mathrm{=}|\mathcal{C}|\mathrm{=}36$) and labeled with the corresponding dataset label. 
Accordingly, a Support Vector Machine (SVM) is trained and evaluated, see Table~\ref{tab:unsupervised_comparison}.
\begin{table}%
 \small
\caption{Unsupervised concept selection: testing set (5 repetitions)}
\label{tab:unsupervised_comparison}
\centering
\footnotesize
\begin{tabular}{p{2.55cm}|p{0.28cm}p{0.2cm}p{0.2cm}p{0.5cm}p{0.22cm}p{1.0cm}p{0.6cm}}
Label: & sack & can & box & teddy & ball & amphora & plate\\ 
\hline
Mean error~(\%): &4.2& 6.5& 2.5& 8.8& 0& 10.4& 0\\
\end{tabular}
\begin{flushleft}
\end{flushleft}
\end{table}
Discriminative results have been obtained, which allow to assess how reasonable the extracted concepts are, e.g., shapes like \emph{ball}, \emph{plate} or \emph{box} show low cross-validation error, whereas appearance variety of categories that include deformability or strong viewpoint dependability, e.g., \emph{teddy} or \emph{amphora} can appear more ambiguous.

\subsection{Generalization to Other Real-World Datasets}
\label{sec:alter_db}
The following experiment evaluates the generalization capability of the proposed approach.
First, $\mathcal{HE}$ is trained once with the training set of the OSCD dataset.
This training process is unsupervised, i.e., $\mathcal{HE}$ is solely trained with instances in a label-agnostic manner. 
Then, instances from the OSCD dataset are propagated through $\mathcal{HE}$ (see Sec.~\ref{sec:spatial_topo_analysis}).
Based on the resulting stimuli vector of the propagation, concepts $C$
are generated (see Sec.~\ref{sec:desc_topo_analysis}).

\begin{figure}[htbp]
	\small
	\centering  
		\def \scaleImg{1.0} 
		\subfigure[]{\label{fig:eval:instance_variety0}\scalebox{\scaleImg}{\includegraphics[height=0.2\linewidth]{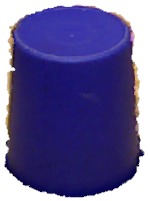}}} \hspace{5pt}
		\subfigure[]{\label{fig:eval:instance_variety1}\scalebox{\scaleImg}{\includegraphics[height=0.17\linewidth]{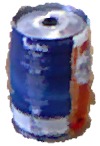}}}  \hspace{5pt}
		\subfigure[]{\label{fig:eval:instance_variety2}\scalebox{\scaleImg}{\includegraphics[height=0.06\linewidth]{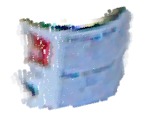}}}  \hspace{5pt}
		\subfigure[]{\label{fig:eval:instance_variety3}\scalebox{\scaleImg}{\includegraphics[height=0.09\linewidth]{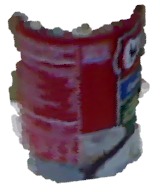}}}  \hspace{5pt}
		\subfigure[]{\label{fig:eval:instance_variety4}\scalebox{\scaleImg}{\includegraphics[height=0.11\linewidth]{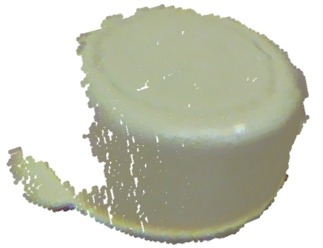}}}  \hspace{5pt}
		\subfigure[]{\label{fig:eval:instance_variety5}\scalebox{\scaleImg}{\includegraphics[height=0.12\linewidth]{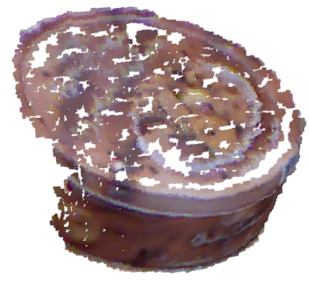}}}
	\caption{Examples of appearance variations of sample point clouds related to the concept \emph{can}, respectively cylinder, from different distributions (datasets): \subref{fig:eval:instance_variety0}, \subref{fig:eval:instance_variety1} show \emph{can}~\emph{0} and \emph{56} of OSCD-training set, \subref{fig:eval:instance_variety2}, \subref{fig:eval:instance_variety3} show \emph{food\_can\_1\_1\_1} and \emph{food\_can\_14\_1\_1} of \emph{WD} and \subref{fig:eval:instance_variety4}, \subref{fig:eval:instance_variety5} show cylindrical instances from scenes \emph{learn 34} and \emph{test 42} of  \emph{SD}.}
	\label{fig:eval:instance_variety}
\end{figure}

Given the previously trained $\mathcal{HE}$ model and the generated concepts $C$, 
we evaluate in the following the discriminative power of the concepts with instances from different real-world datasets. In addition to the OSCD objects, additional datasets with completely different real-world objects are used, namely the \emph{Washington RGB-D Object Dataset} \cite{5980382} (WD) and the \emph{Object Segmentation Database} \cite{6385661} (SD) (see Table~\ref{fig:tsne_instances_dist}); note that all three datasets are sampled from \emph{different distributions} as illustrated in Fig.~\ref{fig:eval:instance_variety0}-\subref{fig:eval:instance_variety5}.

In order to analyze the spectrum of responses for these dataset objects, each object $o$ is initially represented with as graph of segments $g^o$ (see Sec.\ref{sec:obj_seg_dict}) and applied to the two-step procedure: \textbf{1)} propagate $g^o$ through $\mathcal{HE}$ to generate a stimuli vector  $\leftidx{^*}\gamma^o$ (see Sec.~\ref{sec:stimuli_generation}); \textbf{2)} compute for each concept $c\in \mathcal{C}$ the response with $\phi^c(\leftidx{^*}\gamma^o)$ (see Eq.~\ref{eq:concept_response} in Sec.~\ref{sec:shape_concept_inference}).
As a result, an object $o$ generates a set of concept responses $\rho^o\mathrm{=}\{ \phi^1(\leftidx{^*}\gamma^o),\phi^2(\leftidx{^*}\gamma^o),...\}$ ($|\rho^o|\mathrm{=}|\mathcal{C}|\mathrm{=}36$, see Fig.~\ref{fig:ph_concept_extraction}).

\begin{figure}[tb]
	\footnotesize
	\centering
	\includegraphics[width=0.99\linewidth]{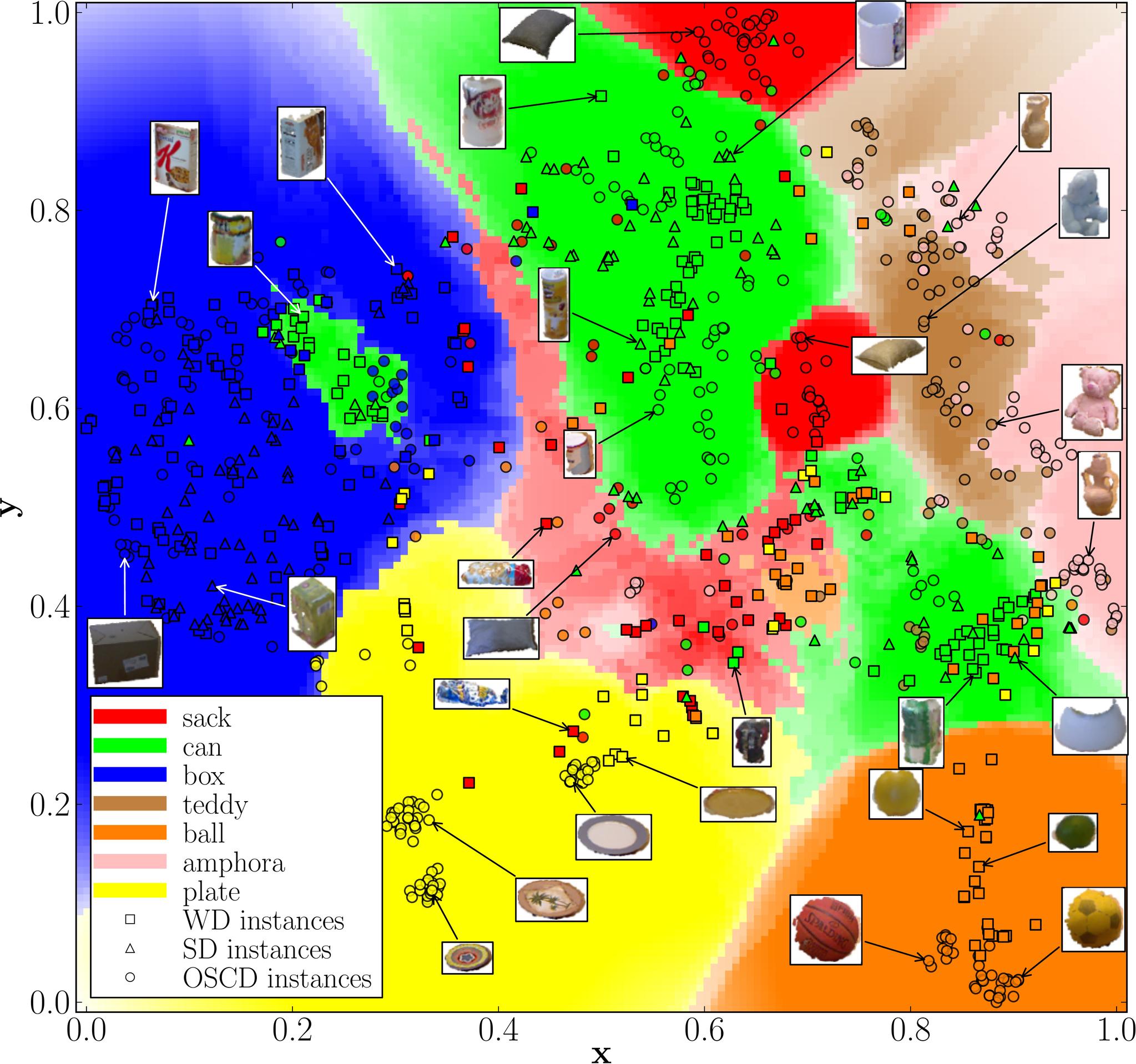}
	\caption{$\mathcal{CR}^2$ with instances from the WD, SD and OSCD datasets (see Table~\ref{fig:tsne_instances_dist}). The instance annotations are scaled for better visibility.}
	\label{fig:unsup_tsne}
\end{figure}

Consequently, a $|\mathcal{C}|$-dimensional space of concept responses $\mathcal{CR}^{|\mathcal{C}|}$ is created.
The generalization capability can be assessed by $\mathcal{CR}^{|\mathcal{C}|}$, which allows to observe relations and similarities among sample objects.
To visualize and reason about the $|\mathcal{C}|$-dimensional $\mathcal{CR}^{|\mathcal{C}|}$ space, the \emph{t-SNE}~\cite{ictdbid:2777} embedding technique is applied to reduce the dimensionality to two; we denote this 2D space as $\mathcal{CR}^2$.
The embedding is performed in an unsupervised manner, i.e., it is label-agnostic.
Fig.~\ref{fig:unsup_tsne} shows instances from the WD, SD and OSCD datasets projected to the two-dimensional $\mathcal{CR}^2$ space.

\begin{table}[tb]
\caption{Sample distribution of the $\mathcal{CR}^2$ space}
\centering
\scriptsize
\setlength{\tabcolsep}{1.0pt}
\renewcommand{\arraystretch}{0.99}
\begin{tabular}[b]{ l|| l |r|l| r |l |r ||r }
	Label& \multicolumn{2}{c|}{WD~\cite{5980382} scans\ \ \#} &  \multicolumn{2}{c|}{SD~\cite{6385661} scans \ \#}&  \multicolumn{2}{c||}{OSCD~\cite{mueller_birk_iros2018} scans\ \ \ \ \ \ \ \ \ \#}&$\Sigma$\\ \hline
	\hline		
	\multirow{2}{*}{\emph{sack}} &  \emph{food bag} 1-8 & 40& & & \emph{sack} 0-56 (tr. set)&57&\multirow{2}{*}{115}\\
	&  & & & &\emph{sack} 0-17 (te. set)&18&\\ \hline
	\multirow{2}{*}{\emph{can}}  &  \emph{food can} 1-14 &70&  \emph{learn} 33-44 &38& \emph{can} 0-59 (tr. set)&60&\multirow{2}{*}{259}\\
	&  \emph{soda can} 1-6 & 30&\emph{test} 31-42&42      & \emph{can} 0-18 (te. set) & 19&\\ \hline
	\multirow{2}{*}{\emph{box}} & \emph{cereal box} 1-5 & 25& \emph{learn} 0-16&38& \emph{box} 0-53  (tr. set)&54&\multirow{2}{*}{232}\\
	& \emph{food box} 1-12 & 60&            \emph{test} 0-15 &36& \emph{box} 0-18 (te. set) &19&\\ \hline
	\multirow{2}{*}{\emph{teddy}} &   &  & & &teddy 0-44 (tr. set) &45 &\multirow{2}{*}{59}\\
	&   &  & & &teddy 0-13 (te. set) &14 &\\ \hline
	\multirow{3}{*}{\emph{ball}}&  \emph{ball} 1-7 & 35&     &&\emph{ball} 0-39  (tr. set)&40&\multirow{3}{*}{125}\\
	&  \emph{lime} 1-4 & 20&                 && \emph{ball} 0-9 (te. set)&10&\\
	&  \emph{orange} 1-4 & 20&                &&  &&\\ \hline
	\multirow{2}{*}{\emph{amphora}}&   &  & & &\emph{amphora} 0-47 (tr. set)& 48&\multirow{2}{*}{62}\\
	&   &  & & &\emph{amphora} 0-13 (te. set)& 14\\ \hline
	\multirow{2}{*}{\emph{plate}}& \emph{plate} 1-7 &35&      && \emph{plate} 0-49 (tr. set)&50&\multirow{2}{*}{105}\\ 
	& &&      && \emph{plate} 0-19 (te. set)&20 \\ \hline \hline
	$\Sigma$&  \multicolumn{1}{c|}{-} & 335 &     \multicolumn{1}{c|}{-}      &154&       \multicolumn{1}{c|}{-}  & 468 & 957
\end{tabular}  	 
\label{fig:tsne_instances_dist}
\begin{flushleft}
	Note, for each instance of WD the $1^{st}$ to $5^{th}$ point cloud scans are selected of the first video sequence. (tr.$=$training, te.$=$testing)
\end{flushleft}
\end{table}

\begin{figure}[tb]
	\small
	\centering  
	\subfigure[]{\label{fig:eval:oscd_cfmat:dist}\includegraphics[width=0.49\linewidth]{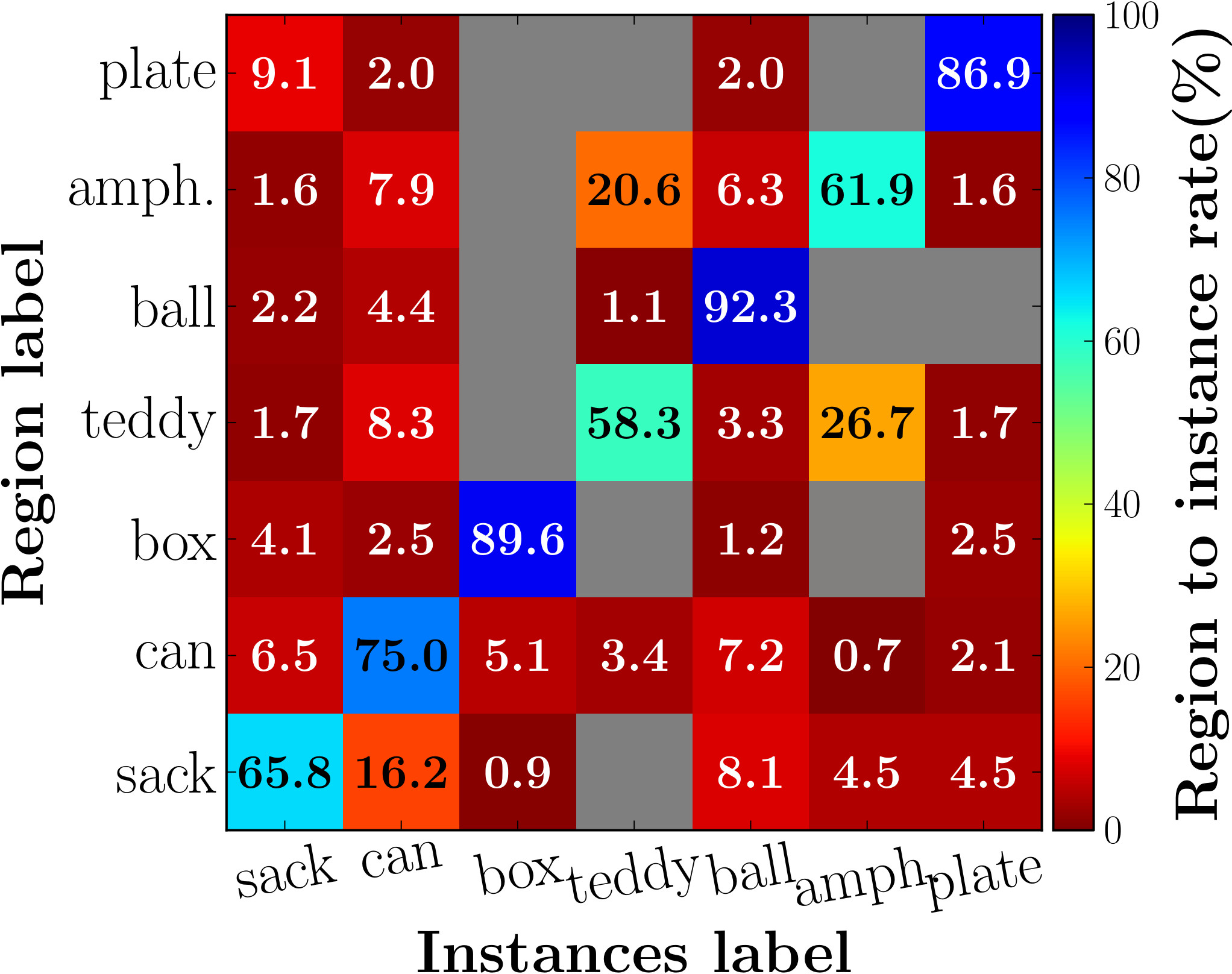}}
	\subfigure[]{\label{fig:eval:oscd_cfmat:assigm}\includegraphics[width=0.4825\linewidth]{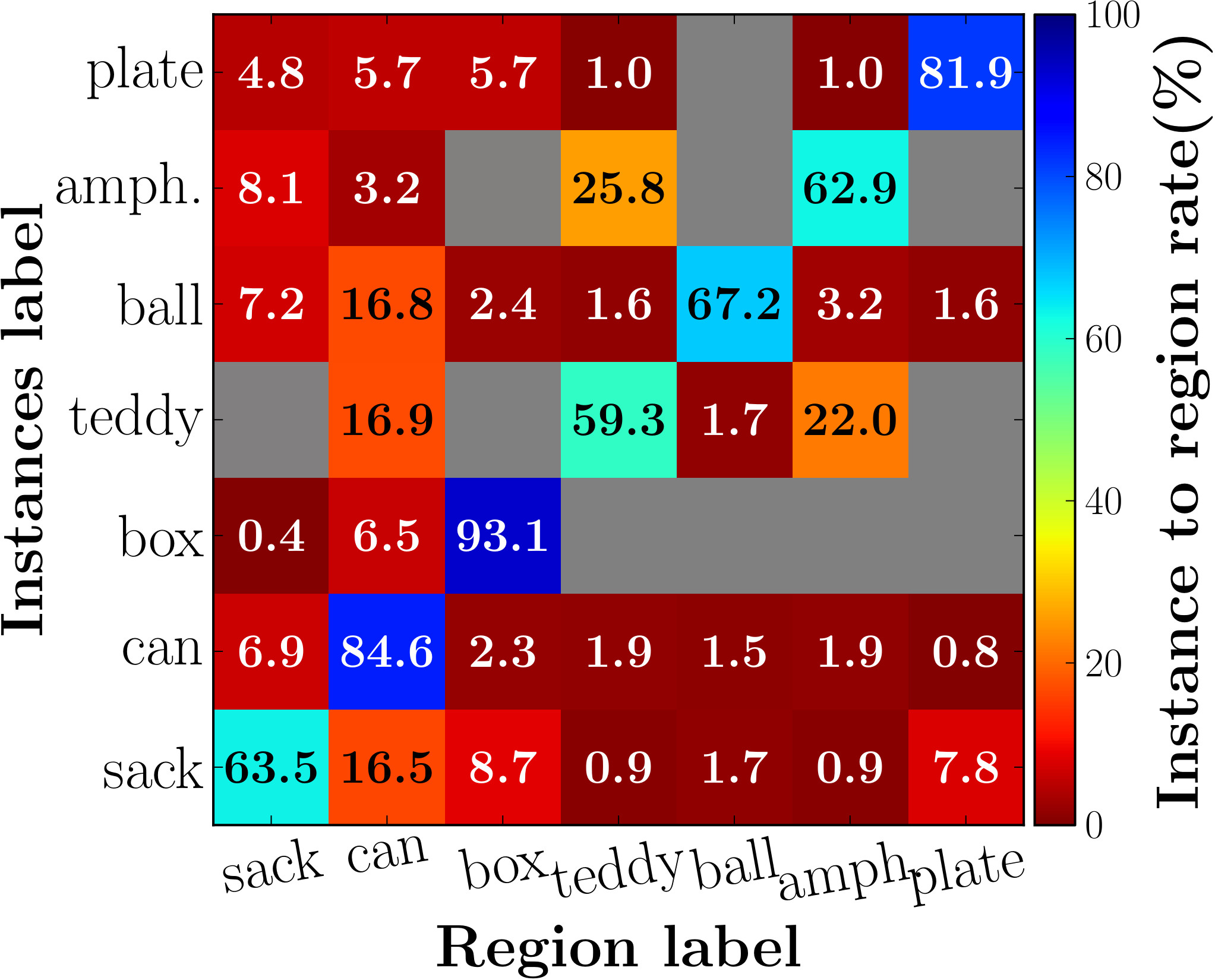}}
	\caption{According to $\mathcal{CR}$ in Fig.~\ref{fig:unsup_tsne}, the distribution is shown of instances within a region \subref{fig:eval:oscd_cfmat:dist} and assignment of instances to particular regions \subref{fig:eval:oscd_cfmat:assigm}.
	}
	\label{fig:eval:oscd_cfmat}
\end{figure}
For illustration, regions in $\mathcal{CR}^2$ are colored according to their correlation with a certain label (see Fig.~\ref{fig:unsup_tsne}) by exploiting the projected instances as anchor points in space. 
A uniform grid is created in the 2D $\mathcal{CR}^2$ space; for each cell in the grid the \emph{k}-nearest instances are determined (e.g., \emph{k}$\mathrm{=}$5\% of total number of instances); then the majority label of the \emph{k} instances is determined and the cell is colored according to the majority label; each
cell is weighted and visually depicted in form of
cell opacity. 
The weight represents the observed proportion of the \emph{k} instances associated to the majority label, which is depicted in an interval $[0, 1]$ from low to high proportion [low: transparent (white)$\mathrm{=}$0, high: opaque (solid majority label color)$\mathrm{=}$1].

The continuous space $\mathcal{CR}^2$ shown in Fig.~\ref{fig:unsup_tsne} allows to observe regional characteristics and relations among locations in $\mathcal{CR}^2$ and instances of the three datasets.
A main observation is that instances from different datasets are propagated through the $\mathcal{HE}$ and the resulting concept responses show a strong coherency with respect to shape appearance: different instances from the different datasets that can be considered to be similar on a human semantic level, form interrelated and coherent groups, as shown by the uniformly colored regions in Fig.~\ref{fig:unsup_tsne}.
This is also reflected in Fig.~\ref{fig:eval:oscd_cfmat:dist} and \subref{fig:eval:oscd_cfmat:assigm} that illustrate the distribution of instances in $\mathcal{CR}^2$ space.
Instances labeled as \emph{can}, \emph{box}, \emph{ball}, \emph{amphora}, \emph{plate} form distinct regions whereas deformable instances like \emph{sack} and \emph{teddy} lead to more scatter. 
However, \emph{teddies} are still represented as a connected region and regions dedicated to \emph{sack} are located at transitions to other labeled regions, e.g., \emph{can} to \emph{plate}, \emph{can} to \emph{box} or \emph{can} to \emph{teddy}.
This observation can be explained that \emph{sacks} can be interpreted as an intermediate shape, e.g., between a \emph{box} and a \emph{can} in $\mathcal{CR}^2$ space due to their roundish, bulgy or cylindric appearance depending on viewpoint and deformation.

\section{Experiments: Mental Simulation}
\label{sec:simu:experiment}
In this section, the performance is evaluated when using the mental simulation for training (Sec.\ref{sec:machine_centric_concepts}). To allow a comparison of our approach with other work, the concrete random samples that are used in this experiment are provided as open dataset, which is denoted in the following as \emph{Artificial Object Dataset}\footnote{http://www.robotics.jacobs-university.de/datasets/2018-artificial-object-dataset-v01/index.php} (AOD). Examples of samples of the artificial abstract objects from this dataset are shown in Fig.~\ref{fig:demo:unsup_sim:samples}.
The dataset contains $250$ training samples, which were artificially generated with an equally distributed number of samples per prototype order ($1$ to $5$).
These artificially generated samples are used to generate shape concepts including the $\mathcal{HE}$ generation.

\subsection{Shape Knowledge Transfer from Mental Simulation to Real-World Data}
\label{sec:simu:demo:knowlege_transfer_sim_real}

\begin{figure}[tb]
	\centering
	\hfill
	\subfigure[\emph{can} instance]{\label{fig:sim_real_exp:sim_train_1}\includegraphics[width=0.32\linewidth]{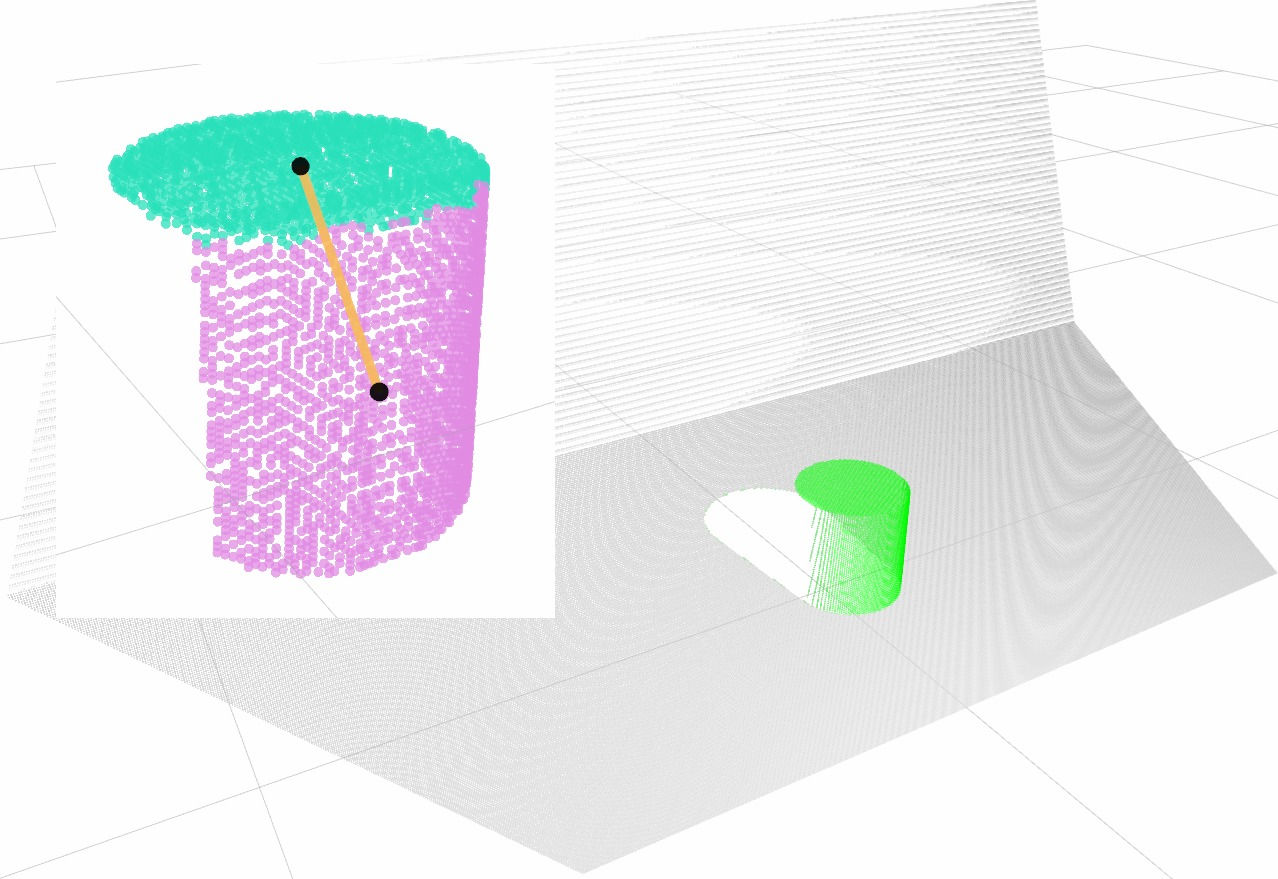}}
	\hfill
	\subfigure[\emph{box} instance]{\label{fig:sim_real_exp:sim_train_2}\includegraphics[width=0.32\linewidth]{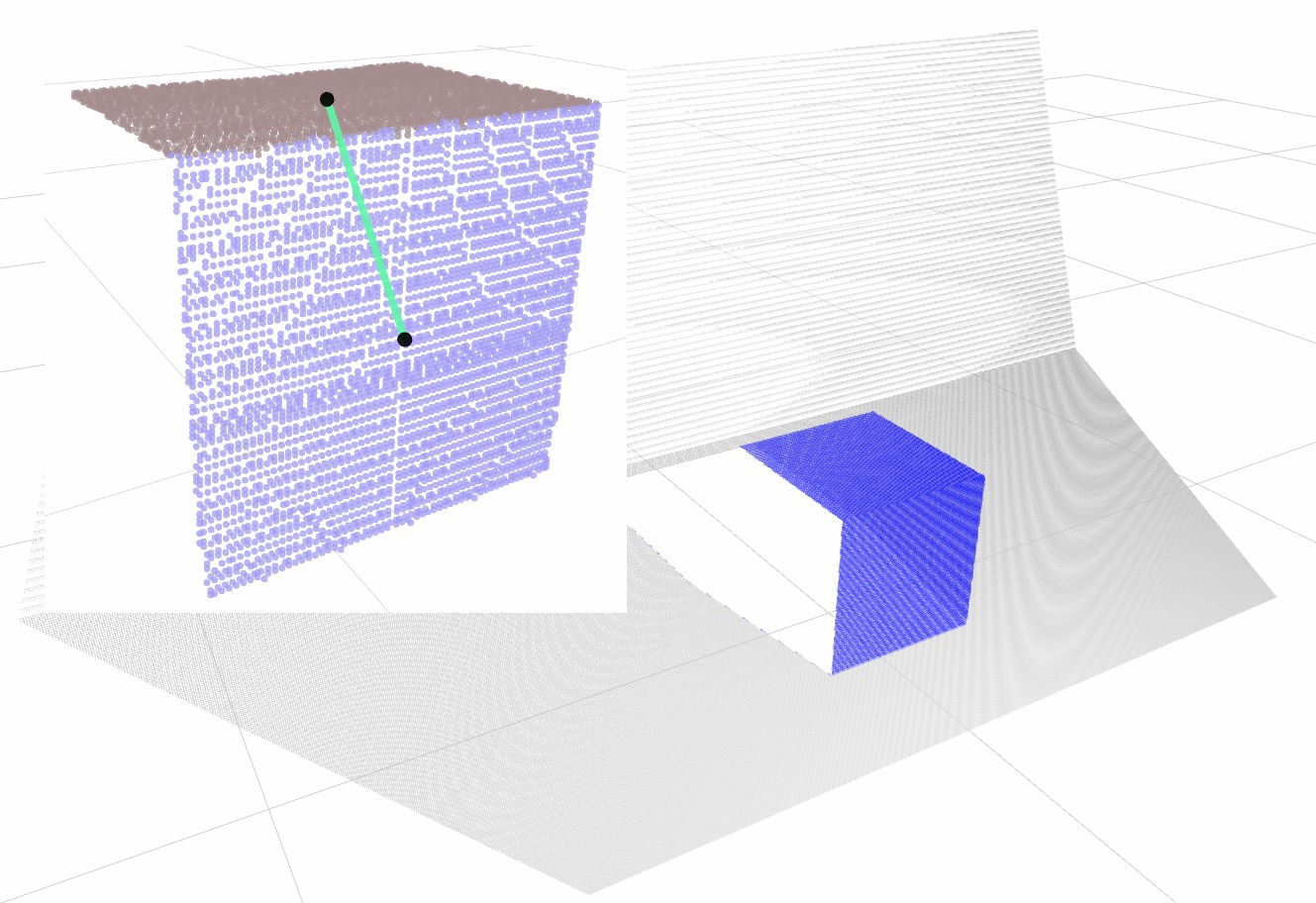}}	
	\hfill
	\subfigure[\emph{ball} instance]{\label{fig:sim_real_exp:sim_train_3}\includegraphics[width=0.32\linewidth]{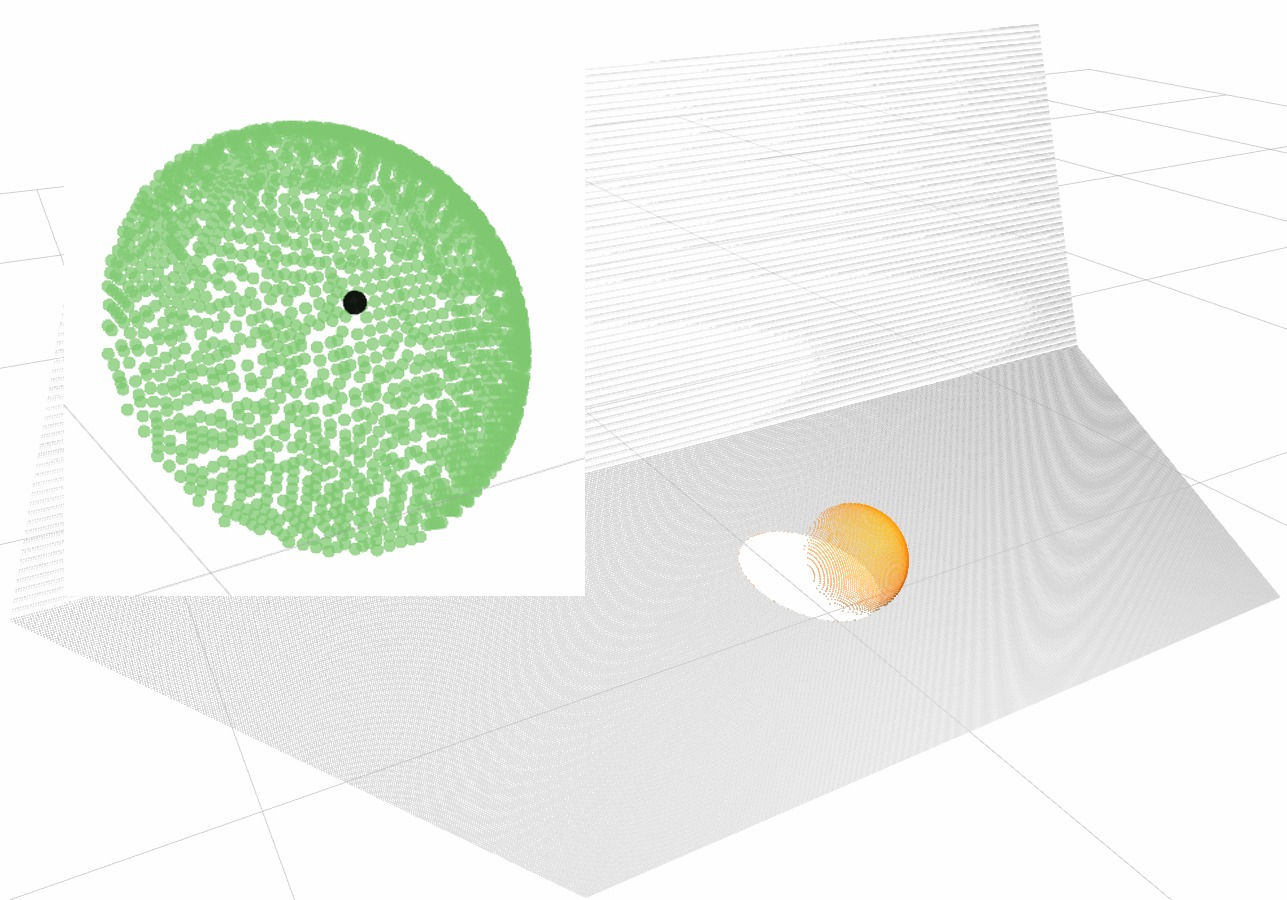}}\hfill
	\caption{Examples of simulated instances of (simple) shape categories and the corresponding extracted super patch graphs (top left in \subref{fig:sim_real_exp:sim_train_1}, \subref{fig:sim_real_exp:sim_train_2}, \subref{fig:sim_real_exp:sim_train_3}). %
	}
	\label{fig:sim_real_exp1}
\end{figure}

We start the evaluation with an illustrative example. 
In Fig.~\ref{fig:sim_real_exp1} three (very simple) simulated objects are shown with their respective extracted segment graph ($g^o$). The simulated instances consist of noise-free point clouds; thus segments are optimally segmented.
When using simulation-based training sample generation, an open question is whether the perception system is able to transfer the knowledge observed in simulation to real object observations.
To test this in this simple illustrative example, the three artificial instances in Fig.~\ref{fig:sim_real_exp1}
are consecutively fed (from Fig.~\ref{fig:sim_real_exp:sim_train_1} to \subref{fig:sim_real_exp:sim_train_3}) 
to $\mathcal{HE}$ and the learned motif prototypes 
are labeled with the respective label.

\begin{figure}[tb]
	\centering
	\centering
	\subfigure[RGB image]{\label{fig:simu:sim_real_exp:real_rgb}\includegraphics[width=0.45\linewidth]{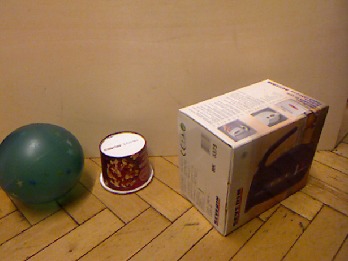}}
	\subfigure[Classification result]{\label{fig:simu:sim_real_exp:real_result}\includegraphics[width=0.5\linewidth]{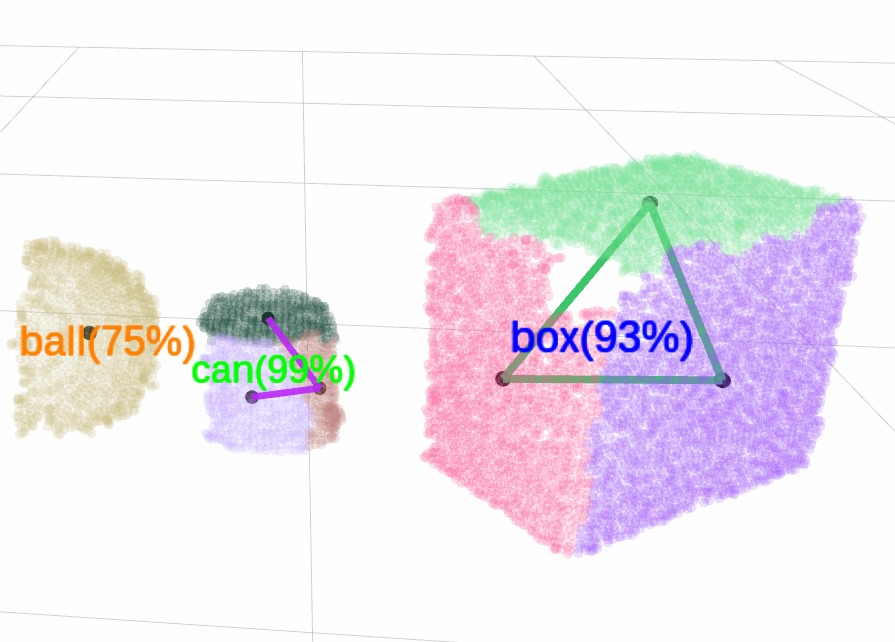}}
	\caption{A sample classification result on real-world scene data using a model trained with simulated objects only (Fig.~\ref{fig:sim_real_exp1}). Note that, objects are segmented from the scene with our previous work~\cite{MuellerBirkIcra2016} and then classified with the trained model.}
	\label{fig:simu:sim_real_exp2}
\end{figure}

In Fig.~\ref{fig:simu:sim_real_exp:real_result} the classification results of this illustrative example on a real-world scene is shown. More precisely, the label with the highest accumulated stimulus considering the observed (labeled) motif prototypes (Sec.~\ref{sec:stimuli_generation}) is shown for each object. 
Note that, $\mathcal{HE}$ has been only trained with a single artificial instance per label (\emph{can}, \emph{box} and \emph{ball}).
Several observation can be made from the classification results.
Considering the correct classification, one may interpret it as a knowledge transfer from simulated data to real noisy observed data.
Regarding sensor noise, segmented surfaces are distorted and may even contain holes (Fig.~\ref{fig:simu:sim_real_exp:real_result}).
These distortions lead to segment constellations, which have not been observed in the training phase. The simulated cylinder (\emph{can}) in Fig.~\ref{fig:sim_real_exp:sim_train_1} naturally consists of an upper planar segment and a cylindric body, whereas the real \emph{can} shown in Fig.~\ref{fig:simu:sim_real_exp:real_result} is over-segmented and subsequently consists of three segments caused by sensor noise. 
Nevertheless, this segment constellation has not been observed in training phase but still led to a correct classification as it is closest to the ideal cylinder concept.
Further on, different viewpoints on objects can lead to different segment constellations due to self-occlusion effects. The viewpoint on the box in Fig.~\ref{fig:sim_real_exp:sim_train_2} results to two planar segments whereas the viewpoint on the box shown in Fig.~\ref{fig:simu:sim_real_exp:real_result} leads to three segments and a hole (red colored segment) caused by sensor noise.
Also in this case, this segment constellation has not been observed in the training phase but it still leads to a correct and confident classification.
Furthermore, note that the simulated instances used for training have in addition completely different spatial dimensions compared to the real objects shown in Fig.~\ref{fig:simu:sim_real_exp:real_result}.

\subsection{Generalization to Real-World Datasets}
\label{sec:gen:simu_dataset}

This experiment evaluates the generalization ability using extensive artificial training data from mental simulation, i.e.,  the Artificial Object Dataset (AOD) with 250 simulated samples of abstract objects.
Initially $\mathcal{HE}$ is trained and concepts are generated once in an unsupervised and label-agnostic manner with the artificial samples of the AOD dataset.
Given the $\mathcal{HE}$ model and the concepts $\mathcal{C}$ generated with the AOD, the generalization capability of $\mathcal{C}$ is evaluated with real-world instances from the \emph{Object Shape Category Dataset} \cite{mueller_birk_iros2018} (OSCD), the \emph{Washington RGB-D Object Dataset}~\cite{5980382} (WD) and the \emph{Object Segmentation Database}~\cite{6385661} (SD), see Table~\ref{fig:tsne_instances_dist}. 
Note that all three real-world datasets are sampled from different distributions (see Fig.~\ref{fig:eval:instance_variety}), i.e., the datasets consist of various, very different objects and they differ with respect to the experimental setups for the sensor data generation.
In order to analyze the spectrum of responses for these dataset objects, each object $o$ is applied to a two-step procedure: \textbf{1)} propagate $o$ through $\mathcal{HE}$ to generate a stimuli vector  $\leftidx{^*}\gamma^o$ (see Sec.~\ref{sec:simu:desc_rep}); \textbf{2)} compute for each concept $c\in C$ the response with $\phi^c(\leftidx{^*}\gamma^o)$.
As a result, an object $o$ is represented by the set of concept responses $\rho^o\mathrm{=}\{ \phi^1(\leftidx{^*}\gamma^o),\phi^2(\leftidx{^*}\gamma^o),...\}$ ($|\rho^o|\mathrm{=}|\mathcal{C}|\mathrm{=}28$).

\begin{figure}[tb]
	\footnotesize
	\centering
	\includegraphics[width=0.99\linewidth]{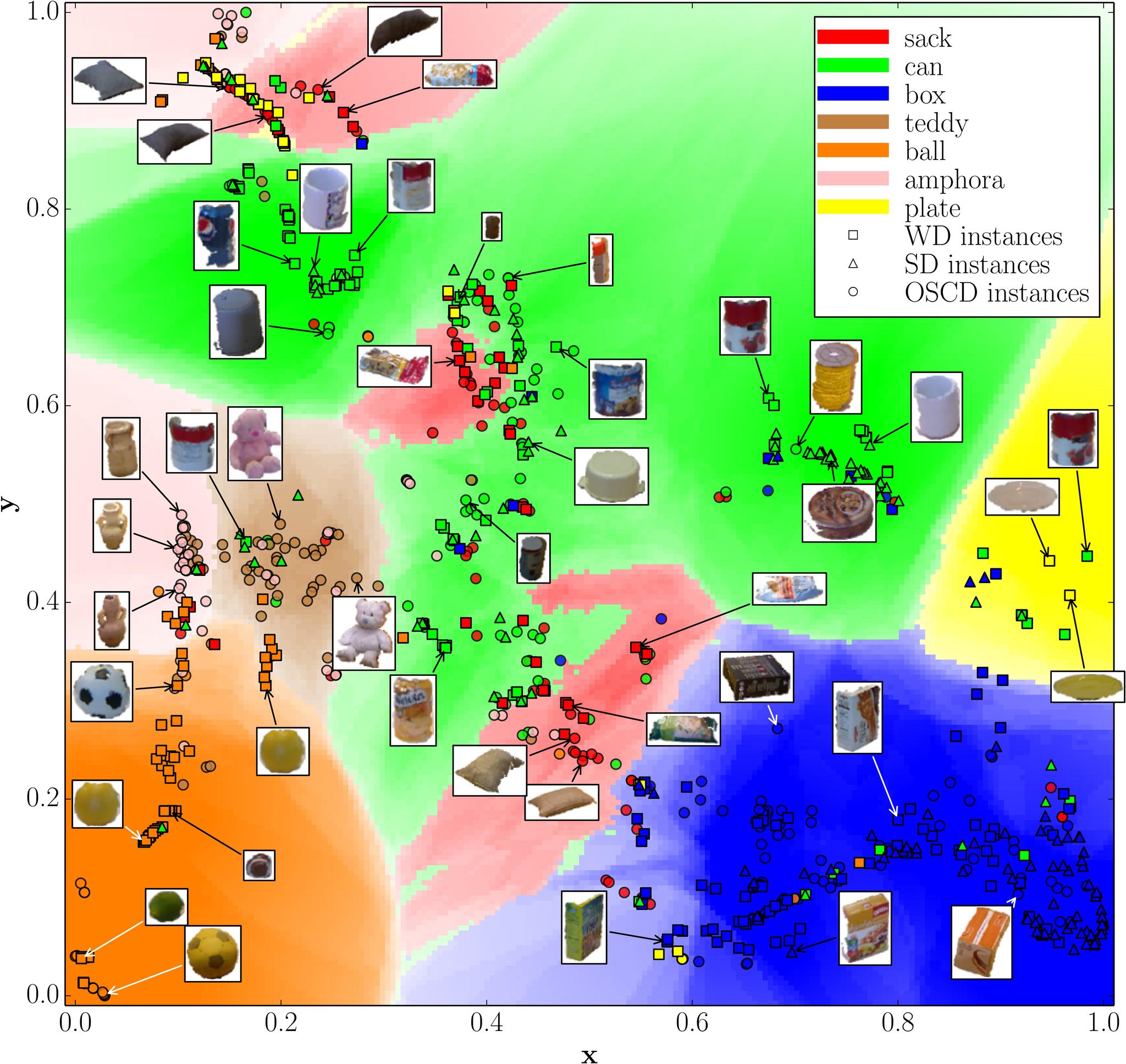}
	\caption{The projection of real-world samples from the OSCD, the WD and the SD dataset to the $\mathcal{CR}$ space that is generated with mental simulation (Fig.~\protect\ref{fig:demo:unsup_sim:samples}). A summary of the instances used is shown in Table~\ref{fig:tsne_instances_dist}.
	}
	\label{fig:demo:unsup_sim:tsne}
\end{figure}

Consequently, in order to investigate the generalization capability, the approach as described in Sec.~\ref{sec:alter_db} is followed, i.e., a $|\mathcal{C}|$-dimensional space of concept responses $\mathcal{CR}^{|\mathcal{C}|}$ is created and the embedding is performed to reduce the dimensionality to two.
As a result, instances from the WD, the SD and the OSCD datasets are projected to this two-dimensional $\mathcal{CR}^2$ space (Fig.~\ref{fig:demo:unsup_sim:tsne}).

\begin{figure}[tb]
	\centering
	\subfigure[]{\label{fig:unsup_sim:reg_ins}\includegraphics[width=0.4925\linewidth]{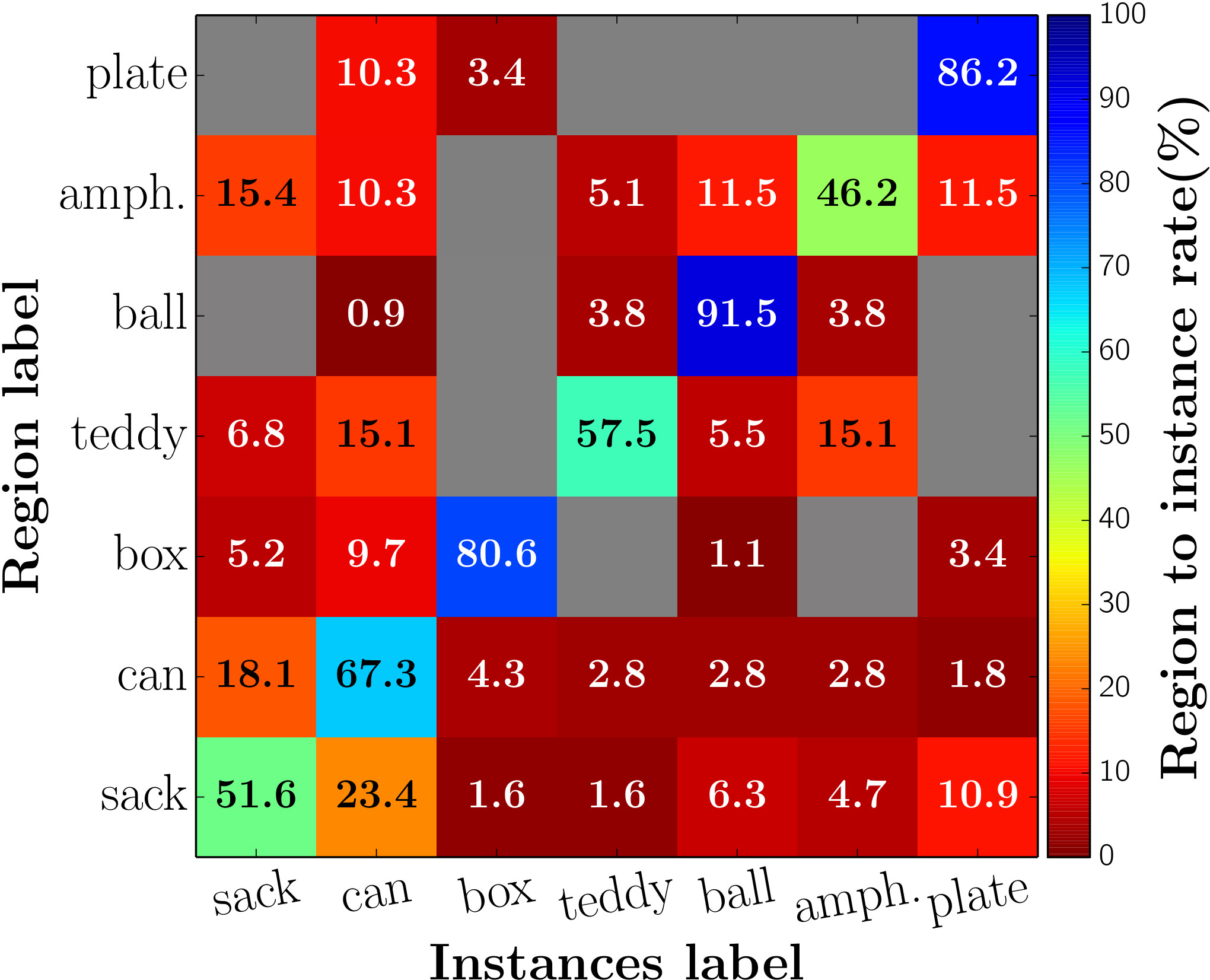}}	\hfill
	\subfigure[]{\label{fig:unsup_sim:ins_reg}\includegraphics[width=0.4925\linewidth]{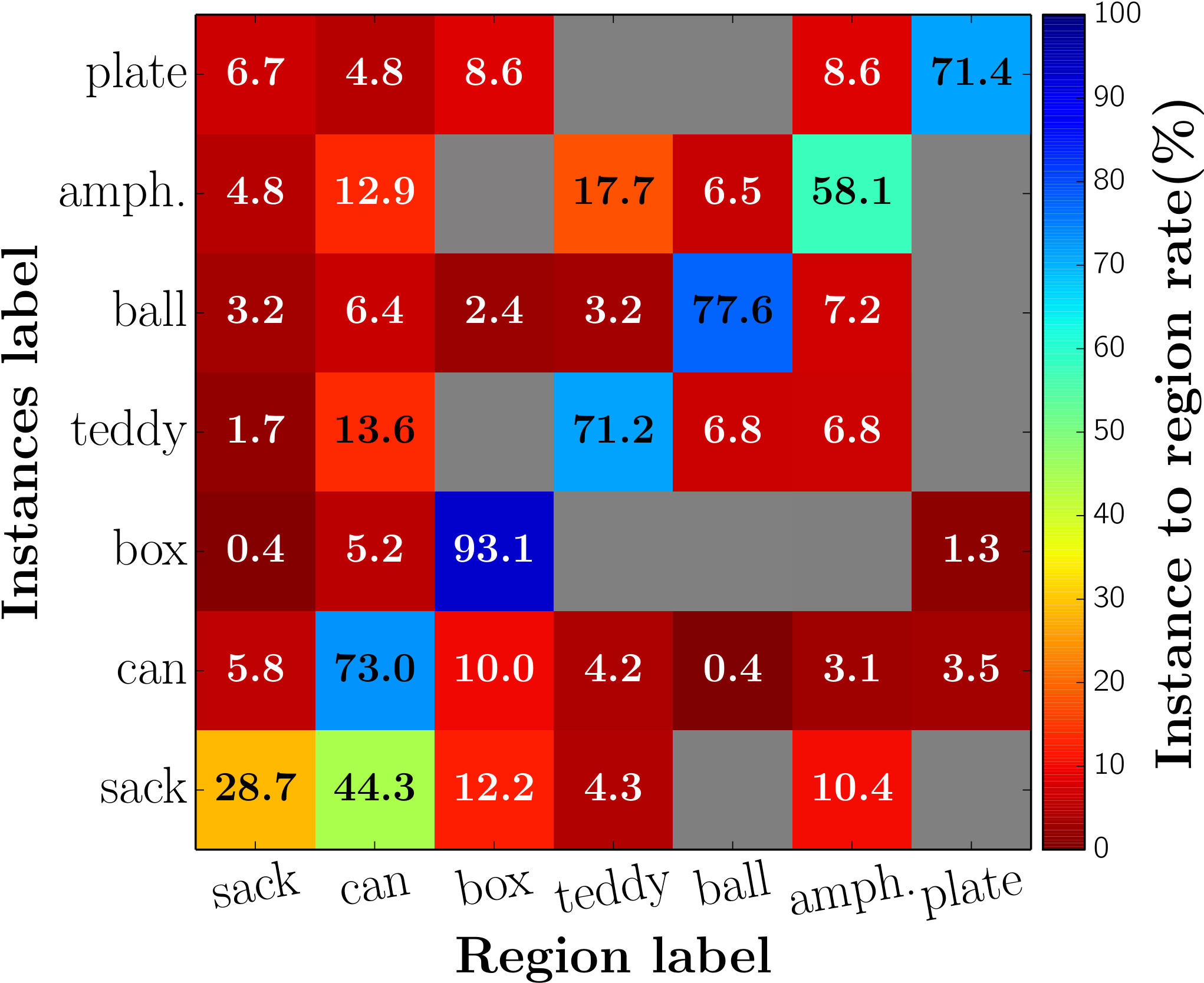}}
	\caption{
		According to $\mathcal{CR}$ in Fig.~\ref{fig:demo:unsup_sim:tsne}, the distribution is shown of instances within \emph{a region} \subref{fig:unsup_sim:reg_ins} and assignment of instances to \emph{particular regions} \subref{fig:unsup_sim:ins_reg}.
	}
	\label{fig:unsup_sim:cf_mat}
\end{figure}

When looking at $\mathcal{CR}^2$, an important observation is that 
after propagating the instances of the three datasets through the $\mathcal{HE}$, the resulting concept responses show also here coherency regarding shape appearance as in shown in Sec.~\ref{sec:alter_db}. Instances of all evaluated datasets together form interrelated and coherent groups, see in Fig.~\ref{fig:demo:unsup_sim:tsne} uniformly colored regions according to the labels of the real datasets.
This is also reflected in Fig.~\ref{fig:unsup_sim:cf_mat} illustrating the instance distribution in $\mathcal{CR}^2$ space. 
By averaging the diagonal (bottom-left to top-right) one can observe $68.7\%$ (Fig.~\ref{fig:unsup_sim:reg_ins}) / $67.6\%$ (Fig.~\ref{fig:unsup_sim:ins_reg}) vs. $75.7\%$ (Fig.~\ref{fig:eval:oscd_cfmat:dist}) / $73.2\%$ (Fig.~\ref{fig:eval:oscd_cfmat:assigm}), i.e., a similar discrimination has been achieved with the artificially generated training set (Fig.~\ref{fig:unsup_sim:cf_mat}) compared to a real object training set (Fig.~\ref{fig:eval:oscd_cfmat}). Note that in an unsupervised manner $\mathcal{CR}$ forms regions of various shapes and degree of label-association in a continuous space (Fig.~\ref{fig:demo:unsup_sim:tsne}) compared to these hard-assigned \emph{discrete} results w.r.t. labels in Fig.~\ref{fig:eval:oscd_cfmat} and \ref{fig:unsup_sim:cf_mat}, which may also contain noise in point clouds and in the labeling process.
Thus, the discrete results may only partially reflect the underlying label-association strength of objects compared to the continuous $\mathcal{CR}$ space.

Consequently, this indicates that randomly generated, abstract instances based on composition of primitive shape prototypes from mental simulation carry information about facets of shape appearance that allow to create shape concepts which facilitate the generation of an abstract space suited to discriminate and categorize real object observations in a reasonable way.
From the perspective of Cognitive Science, specifically in the field of representation architectures, $\mathcal{CR}$ can be seen as a \emph{Conceptual Space}~\cite{2000:CSG:518647,zenker2015,RamaFiorini2014} where points (prototypes) in the abstract space represent multidimensional vectors of \emph{stimuli} and regions in space \emph{concepts}.
These stimuli are often denoted as \emph{Quality Dimensions} and can be interpreted as concept responses $\rho^o$ with respect to $\mathcal{C}$ given an object $o$. 
Another property can be observed that supports that concept responses of similar instances appear close in $\mathcal{CR}$ in comparison to dissimilar ones: the majority of instances of the respective label given by humans are closest or within the same region and form groups (Fig.~\ref{fig:demo:unsup_sim:tsne} and Fig.~\ref{fig:unsup_tsne}).
\section{Conclusion}
\label{sec:conclusion}
We presented an unsupervised abstraction process for machine learning of shape concepts: from 3D point clouds over hierarchically organized motifs to (semantically meaningful) concepts of shape commonalities.
The proposed Shape Motif Hierarchy Ensemble encodes object segment compositions in a hierarchical symbolic manner.
Inspired by the concept of Persistent Homology, stimuli generated by the ensemble are filtered in a gradual manner to reveal topological structures.
The filtration leads to stimuli groups which can be interpreted as shape concepts that reflect commonalities of shape appearances.

An important question is how this unsupervised learning is trained. Even when not using human labels, biases can be in the selection of the dataset instances used for training. Moreover, the generation of real-world datasets is cumbersome and generally requires substantial effort. Therefore, the use of mental simulation is investigated in this article, i.e., the generation of virtual sensor data from artificial abstract objects. 
This approach is unsupervised in two respects: it is label-agnostic (no label information is used) and instance-agnostic (no instances preselected by human supervision are used). 

In a first set of experiments, the shape concepts are learned in an unsupervised, label-agnostic fashion from a single real-world dataset and it is shown that a) semantically meaningful categories emerge, i.e., associations to shape categories linked to human-annotated labels appear, and that b) the concepts generalize to other real-world datasets, i.e., the concepts learned on one dataset lead to meaningful label associations when being applied to completely different real-world datasets. 
In a second set of experiments, these results are extended to mental simulation, i.e., the training is both label-agnostic and instance-agnostic. It is shown that training with virtual sensor data from artificial abstract objects leads to a semantically meaningful shape concept space, which generalizes to real-world object datasets. I.e., it leads to a shape concept space, in which unknown objects of real-world sensor data are grouped (based on their commonalities) into regions in concept space that can be, for instance, linked to human-annotated labels.

\bibliographystyle{IEEEtran}

\end{document}